\documentclass{article} %
\usepackage{iclr2027_conference,times}
\usepackage{graphicx}
\usepackage{float}
\makeatletter
\def\input@path{{./}}
\makeatother
\graphicspath{{./}}
\usepackage{booktabs}
\usepackage{multicol}
\usepackage{multirow}

\usepackage{amsmath,amsfonts,bm}

\def\eqref#1{equation~\ref{#1}}

\def\1{\bm{1}}

\DeclareMathAlphabet{\mathsfit}{\encodingdefault}{\sfdefault}{m}{sl}
\SetMathAlphabet{\mathsfit}{bold}{\encodingdefault}{\sfdefault}{bx}{n}

\usepackage{xcolor}
\usepackage{tcolorbox}
\tcbuselibrary{listings,breakable}
\usepackage{hyperref}
\usepackage{url}
\usepackage{needspace}
\usepackage{fontawesome5}
\usepackage{placeins}
\usepackage{needspace} %
\usepackage{wrapfig}
\makeatletter
\newcommand{\localcaptionof}[1]{\def\@captype{#1}\caption}
\makeatother

\usepackage[nameinlink,capitalize,noabbrev]{cleveref}
\definecolor{tvrlblue}{RGB}{33,91,166}
\hypersetup{colorlinks=true,linkcolor=tvrlblue,citecolor=tvrlblue,urlcolor=tvrlblue}

\providecommand{\suppref}[1]{\cref{#1}}
\providecommand{\mainref}[1]{\cref{#1}}
\newcommand{\Suppref}[1]{\suppref{#1}}

\title{Token-Level Video Reinforcement Learning}

\iclrfinalcopy

\author{Yifan Wang$^{1}$ \quad
Gordon Guocheng Qian$^{\dagger}$ \quad
Yanyu Li \quad
Anil Kag \quad
Yun Fu$^{1}$\\
$^{1}$Northeastern University\\[2pt]
\faGlobe\,\href{https://yfwang1999.github.io/TVRL/}{Project Page}\qquad
\faGithub\,\href{https://github.com/YFWang1999/TVRL}{GitHub}
}

\newcommand{\methodname}{TVRL}

\begin{document}

\maketitle
\lhead{Preprint} %
{\renewcommand{\thefootnote}{\fnsymbol{footnote}}\footnotetext[2]{Corresponding author.}}

\vspace{-8mm}
\begin{figure}[H]
\centering
\includegraphics[width=\textwidth]{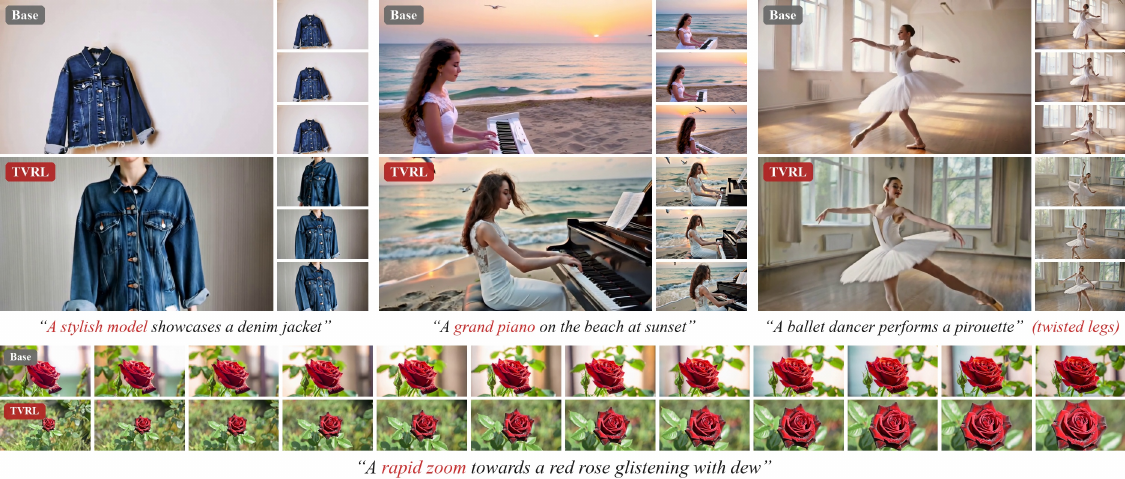}
\caption{{\bf Text-to-video RL post-training with \methodname{}.} Compared with the base model (top rows), \methodname{} (bottom rows) improves subject presence, object fidelity, body plausibility, and camera-motion control; \textcolor{red}{red} marks the prompt phrase to check.}
\label{fig:teaser-results}
\end{figure}

\begin{abstract}
Reinforcement learning (RL) for video generation usually assigns one scalar reward
to an entire sampled video. Yet a video is not uniformly flawed: some visual
tokens may already satisfy the prompt, whereas others require correction.
A scalar reward cannot localize errors, causing optimization
to perturb satisfactory tokens while under-targeting the tokens that actually
need to change. We introduce Token-Level Video Reinforcement Learning, \methodname{}, a framework that derives
token-level credit from the reward being optimized. Our key insight is
that the answer likelihood of a frozen vision-language model provides both
signals: its outputs contribute to the video-level reward, while magnitudes of its
video-input gradients reveal which generated video tokens most affect that score. We
instantiate \methodname{} in Group Relative Policy Optimization by averaging prompt-derived
question rewards into one group-relative advantage and using detached,
question-conditioned token-credit maps to reweight dense denoising-transition
log-probabilities inside the clipped policy ratio. On
VBench-2.0, \methodname{} achieves an Overall score of 57.69, outperforming the
base model by 3.60 points. \methodname{} also improves matched GRPO baselines across three
SDE samplers (SAGE, Flow, and Dance) by 2.68--3.15 points and across four reward
models (VideoAlign, VideoScore2, UnifiedReward2, and Qwen3.5-9B) by 1.33--3.15 points.
\end{abstract}

\section{Introduction}
\label{sec:intro}
\providecommand{\needref}[1]{\textcolor{red}{[REF: #1]}}

Text-to-video generation~\citep{Wan2.1,HunyuanVideo,hunyuan1.5,MovieGen,LTX2} has advanced rapidly with diffusion~\citep{DiffusionModels,DDPM,LDM,VideoDiffusionModels} and flow-matching~\citep{FlowMatching} models.
As base models improve, post-training that aligns generated videos with prompts, human preferences, and evaluation criteria has become increasingly important~\citep{ImageRewardREFL,VideoAlign,VisionReward}.
Inspired by reinforcement learning for language models~\citep{ouyang2022instructgpt,shao2024deepseekmath,deepseekr1}, recent work adapts Group Relative Policy Optimization (GRPO)~\citep{shao2024deepseekmath} to video generation~\citep{Flow-GRPO,DanceGRPO,zheng2026sagegrpo} by converting the deterministic denoising ODE into an SDE sampler, which defines a stochastic policy over latent denoising trajectories.
For each prompt, GRPO samples a group of videos, scores each completed trajectory, and updates the generator with group-relative advantages, with no separate value model.

\begin{wrapfigure}{r}{0.58\columnwidth}
    \vspace{-1.2\baselineskip}
    \centering
    \includegraphics[width=\linewidth]{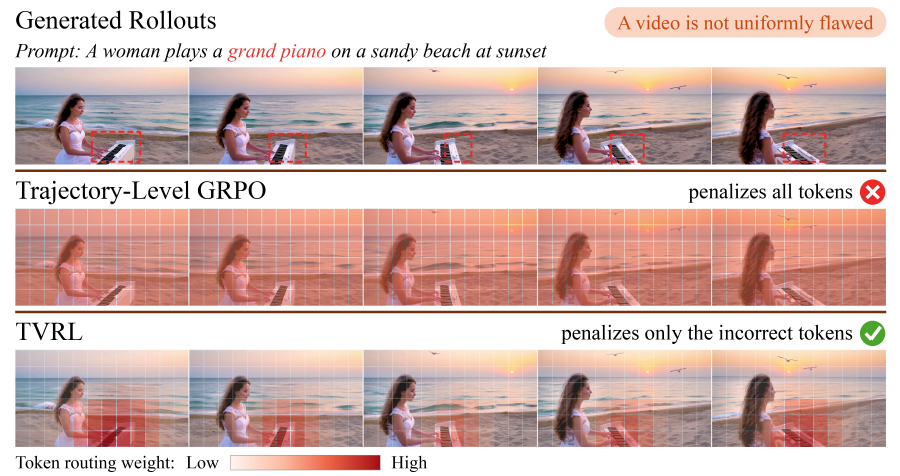}
    \vspace{-5mm}
    \caption{\textbf{Video-level reward, token-level credit.} Video GRPO~\citep{zheng2026sagegrpo} broadcasts one scalar advantage to every video token. \methodname{} provides token-level credit to acknowledge or penalize relevant tokens.}
    \label{fig:illustration}
    \vspace{-2mm}
\end{wrapfigure}

Despite this progress, video RL faces a fundamental credit-assignment bottleneck.
Training must determine not only whether one trajectory is better than another, but also when and where the reward-relevant evidence appears.
A video-level scalar reward can rank completed denoising trajectories, but it cannot indicate which video tokens are responsible for success or failure~\citep{wu2025densedpo,Diffusion-DRF,TeleBoost} (\cref{fig:illustration}).
This limitation matters because the evidence for many prompt requirements is confined to a small region or a few frames, for example, whether the requested grand piano is rendered as a piano rather than a generic keyboard (\cref{fig:teaser-results}), whether a watch case is rose gold, whether the described fireflies appear in the scene at all, or whether the camera pans to the right instead of staying static.
Broadcasting the same scalar advantage to every latent token therefore spends update capacity on regions that have nothing to do with the evaluated error.
Recent work separates the \emph{video-level reward} from \emph{token credit}~\citep{ni2026seeingmattersvisualpreference}, but derives credits from perceptual features independent of rewards.

The missing quantity is a routing signal tied to the reward itself.
Score gradients provide such a signal because they measure how sensitive an evaluator's score is to local changes in the video-frame input~\citep{ImageRewardREFL,DRaFTDiffusionRewardTuning1,AlignPropDiffusionRewardTuning5,VaderVideoDiffusionRewardTuning1}.
Prior differentiable reward methods typically use these gradients to update the generator directly.
In GRPO, we use them differently: after detachment, their magnitudes weight denoising log-probabilities, while the group-relative advantage remains the scalar RL signal.
This preserves the clipped policy-gradient structure of GRPO and avoids backpropagating the reward loss through the generator.
The remaining requirement is semantic alignment: the score being differentiated must represent the requirement whose credit we want to localize.

This score--credit alignment makes the choice of evaluator interface central.
Custom video reward models provide differentiable scalar feedback, but a single scalar head does not expose which semantic requirement its gradient should explain~\citep{Diffusion-DRF}.
We therefore use an evaluator whose score is tied to an explicit video-grounded question, so the reward and the sensitivity map are computed from the same semantic check.
Pretrained VLMs~\citep{qwen3.5,Qwen2.5-VL} offer exactly such an interface: prompt following, physical plausibility, object interaction, and temporal consistency can all be phrased as video-grounded questions, and the teacher-forced likelihood of a reference answer measures whether each check is satisfied.
Because every question--answer likelihood is differentiable with respect to the VLM's video-frame input, a single check provides both a scalar contribution to the rollout objective and a question-conditioned sensitivity map for routing its update.
Using a frozen, off-the-shelf evaluator also removes the need for reward-model training and per-check preference labels~\citep{VLMRewardVideoDiffusionRLTrain3}.
This is the central idea of our work: the same semantic check that determines \emph{whether} a prompt requirement is satisfied also estimates \emph{where} the resulting policy update should be applied.

Building on this observation, we propose \methodname{} (Token-Level Video Reinforcement Learning), a training framework in which a video policy is evaluated by video-level rewards but updated through explicit token-level credit; the name does not assume that video tokens are discrete actions. We instantiate the framework for text-to-video diffusion with a clipped group-relative objective.
For each prompt, we construct binary video-grounded checks and score generated videos by the teacher-forced VLM likelihood of the reference answers.
Each check yields two coupled quantities from the same scalar: a question reward given by the likelihood value, and a detached token-credit proxy given by the magnitude of its video-frame-input gradient. The latter measures local sensitivity rather than signed or causal credit.
We average the question rewards before group-relative normalization to obtain a single rollout-level advantage shared across checks, and we use each question-conditioned credit map to reweight the dense denoising-transition log-probabilities inside the clipped GRPO surrogate.
The shared advantage sets the direction and strength of the update, while the credit maps determine its spatio-temporal allocation.
This design preserves group-relative rollout ranking and ratio clipping, and it never backpropagates reward gradients through the generator.
Averaging several checks also reduces reliance on any single question.
Our \textbf{contributions} are:
\begin{itemize}
    \item We derive token-level credit for video GRPO from the reward itself: a frozen VLM supplies both a question reward and a detached, question-conditioned sensitivity map, requiring neither task-specific reward-model training nor dense credit labels.
    \item We develop a shared-advantage, localized-routing objective that aggregates question rewards into one rollout-level advantage and uses the credit maps to reweight dense denoising-transition log-probabilities within the clipped GRPO ratio. Under the baseline's log-probability reduction, uniform weights recover scalar-reward GRPO.
    \item With the SAGE sampler, \methodname{} improves VBench-2.0 Overall over the corresponding GRPO baselines by +1.36, +1.33, +1.85, and +3.15 points with VideoAlign, VideoScore2, UnifiedReward2, and Qwen3.5-9B rewards, respectively. With the Qwen3.5-9B reward fixed, it also improves over GRPO under the SAGE, Flow, and Dance samplers by +3.15, +2.92, and +2.68 points. Ablations cover credit granularity and the choice of VLM critic.
\end{itemize}

\section{Related Work}
\label{sec:related}

\paragraph{Token-level credit for video optimization.}
Alignment of video generators has increasingly moved beyond a single global score toward feedback structured by space, time, or evaluation criterion.
Differentiable reward optimization obtains such feedback by backpropagating reward gradients through the decoder and the denoising chain~\citep{DRaFTDiffusionRewardTuning1,AlignPropDiffusionRewardTuning5,VaderVideoDiffusionRewardTuning1,InstructVideoVideoDiffusionRewardTuning2,Diffusion-DRF}.
These gradients encode how sensitive the score is to local video evidence, but using them to update the generator directly requires keeping the reward model, the decoder, and the multi-step denoising graph in memory, and repeated Jacobian products along the stochastic chain can amplify unstable local gradients.
Other dense-feedback methods expose localized failures through temporal preference labels, dense captions, rich human or AI feedback, or patch-level reward models~\citep{wu2025densedpo,ActivityNetCaption,LVD-2M-Temporal-Dense-Caption-Dataset,RichHumanFeedbackGoogle,VideoAlign,wang2025harnesslocalrewardsglobal}, at the cost of additional annotation or a separately trained localized reward model.
Structured alignment systems~\citep{TeleBoost} and ViPO~\citep{ni2026seeingmattersvisualpreference} redistribute a scalar evaluation or GRPO advantage to finer units, yet their allocation maps come from perceptual priors rather than from the check that produced the reward.
Recent VLM-based video evaluators instead express prompt requirements as semantic questions rather than task-specific reward heads~\citep{qwen3.5,VideoScore2,wang2025unified}.
\methodname{} uses this interface to couple reward evaluation with token-level credit: teacher-forced answer likelihoods score prompt-level checks, and detached video-frame-input gradients weight the denoising log-probabilities within GRPO.
Unlike Diffusion-DRF~\citep{Diffusion-DRF}, which also uses structured VLM feedback but backpropagates it through the generator, the VLM gradient here stops at the credit map; unlike scalar GRPO, it changes where the update is applied; and unlike dense-feedback methods, it needs no dense labels or localized reward model.

\paragraph{GRPO for video generation.}
GRPO removes the need for a learned value model by sampling multiple outputs for the same prompt and normalizing their rewards within the group, which makes it attractive for large video generators~\citep{shao2024deepseekmath}.
Recent video GRPO methods adapt this formulation to diffusion and flow models by treating denoising as a stochastic policy and defining transition log-probabilities along sampled latent trajectories.
Flow-GRPO~\citep{Flow-GRPO}, DanceGRPO~\citep{DanceGRPO}, and SAGE-GRPO~\citep{zheng2026sagegrpo} instantiate this framework with different stochastic denoising samplers.
All of them optimize a clipped policy-gradient objective, but with a scalar reward they assign one terminal reward to each rollout and apply the resulting group-relative advantage uniformly across denoising steps and latent video tokens.
Such a reward indicates whether a video is better than the others in its prompt group, not where the update should concentrate.
Our video instantiation of \methodname{} is complementary to these samplers: it retains the clipped group-relative update and replaces uniform assignment with VLM-gradient token credit that reweights the dense denoising log-probabilities before the policy ratio is formed.

\section{Method}
\label{sec:method}
\begingroup
\setlength{\abovedisplayskip}{5pt plus 1pt minus 2pt}
\setlength{\belowdisplayskip}{5pt plus 1pt minus 2pt}
\setlength{\abovedisplayshortskip}{3pt plus 1pt minus 1pt}
\setlength{\belowdisplayshortskip}{3pt plus 1pt minus 1pt}
\setlength{\textfloatsep}{9pt plus 2pt minus 2pt}

In \emph{token-level video reinforcement learning}, a video-level reward decides whether a rollout is reinforced or suppressed, and an explicit credit distribution decides how that update is allocated over video tokens. With rollout advantage $A_i$, local policy score $u_{i,t,g}^{\theta}$ of video unit $g$, and credit $w_{i,g}\geq0$ with $\sum_g w_{i,g}=1$, the unclipped surrogate gradient at the old policy is $-A_i\sum_g w_{i,g}\nabla_\theta u_{i,t,g}^{\theta}$. Uniform credit recovers scalar GRPO; ViPO~\citep{ni2026seeingmattersvisualpreference} derives $w$ from perceptual features, whereas we derive it from the reward's own input gradient.

We instantiate this for a text-to-video diffusion policy $\pi_\theta$ with a frozen vision-language model (VLM). Given a prompt $c$, the policy samples a denoising trajectory $\tau=(\mathbf{z}_T,\ldots,\mathbf{z}_0)$, a frozen decoder produces the video $\mathbf{v}=D(\mathbf{z}_0)$, and the VLM scores sampled frames by teacher-forced question answering. The question rewards form one group-relative advantage per rollout, and their detached VLM gradients only route it over latent units inside the clipped GRPO objective; no VLM gradient reaches the generator (\cref{fig:tvrl-pipeline}).

\begin{figure*}[t]
\centering
\includegraphics[width=\textwidth]{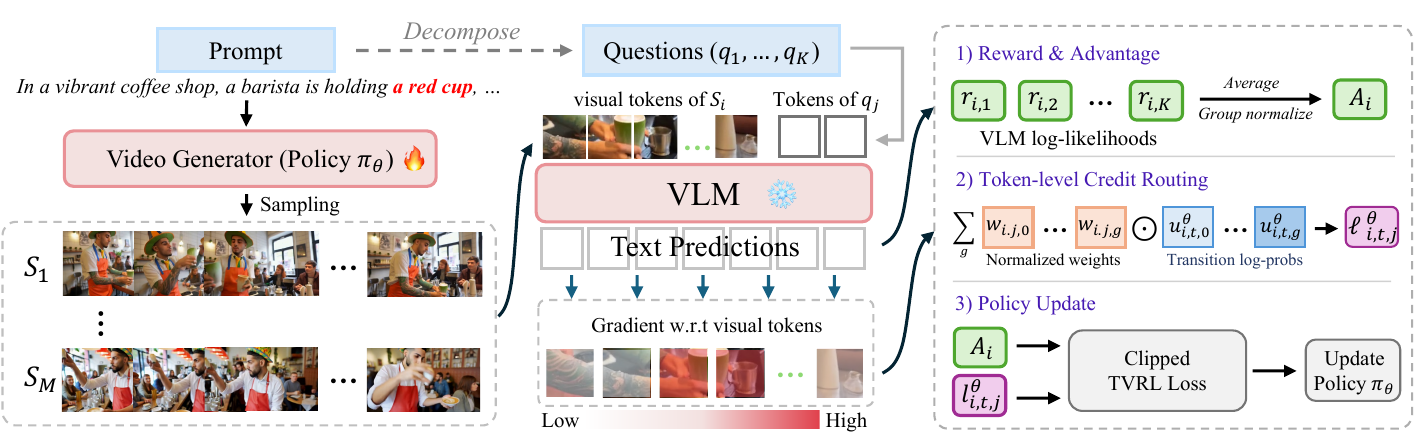}
\caption{Overview of \methodname{}. Teacher-forced likelihoods for $K$ prompt-derived checks are averaged within each of $M$ rollouts and normalized across the rollout group into one advantage per video. Gradients of the same likelihoods yield detached, question-conditioned weights that route dense denoising log-probabilities inside the clipped GRPO update.}
\label{fig:tvrl-pipeline}
\end{figure*}

\subsection{Policy and video-grounded checks}
\label{sec:method_prelim}
\label{sec:method_questions}

A pretrained diffusion transformer~\citep{DiT,Wan2.1} defines stochastic transitions $\pi_\theta(\mathbf{z}_{t-1}\mid\mathbf{z}_t,t,c)$. During GRPO rollouts, the SDE sampler perturbs the transition mean $\boldsymbol{\mu}^{\mathrm{sde}}_\theta$ with noise schedule $\sigma_{\eta,t}$:
\begin{equation}
\begin{aligned}
    \mathbf{z}_{t-1}
    &\sim \mathcal{N}\!\left(
    \boldsymbol{\mu}^{\mathrm{sde}}_\theta(\mathbf{z}_t,t,c),
    \sigma_{\eta,t}^{2}I\right),\\
    \log\pi_\theta(\mathbf{z}_{t-1}\mid\mathbf{z}_t,t,c)
    &=-\frac{\|\mathbf{z}_{t-1}-\boldsymbol{\mu}^{\mathrm{sde}}_\theta(\mathbf{z}_t,t,c)\|_2^2}
    {2\sigma_{\eta,t}^{2}}+\mathrm{const.}
\end{aligned}
\label{eq:main_diffusion_transition}
\end{equation}
This transition yields an elementwise latent log-probability map whose gradients reach the generator through the transition mean. Different SDE samplers change the mean and the noise schedule, but \methodname{} requires only that the dense log-probabilities can be reduced over a chosen set of credit units. \Suppref{app:diffusion_policy_details} gives the full scheduler transition.

We decompose each prompt offline into $K$ atomic yes/no checks $\mathcal{Q}(c)=\{q_j(c)\}_{j=1}^{K}$. The decomposition model sees only the prompt, and each check targets a single requirement directly verifiable from the generated video, such as a subject, attribute, action, relation, scene element, or camera cue. At reward time, all checks and their fixed reference answers are placed in a structured schema and scored in a single teacher-forced VLM pass (templates in \suppref{app:decomposition_prompt,app:vlm_reward_prompt}).

\subsection{Teacher-forced reward and token credit}
\label{sec:method_grad_reward}

Let $p_\phi$ denote the frozen VLM. For rollout $i$ and check $j$, the question reward is the average log-probability of the fixed reference answer over its answer-token positions $\mathcal{A}_{i,j}$:
\begin{equation}
    R_{i,j}
    =
    \frac{1}{|\mathcal{A}_{i,j}|}
    \sum_{\ell\in\mathcal{A}_{i,j}}
    \log p_\phi\!\left(a_{i,j}^{\ell}\mid \mathbf{x}_{i},a_i^{<\ell}\right).
    \label{eq:vlm_question_score}
\end{equation}
We differentiate the same score with respect to the processed video-frame input $\boldsymbol{\xi}_i$. For a credit granularity $\Gamma$ with latent credit units $\mathcal{G}_\Gamma$, the gradient magnitude is aggregated, aligned with these units, and normalized:
\begin{equation}
\begin{aligned}
    \mathbf{g}_{i,j} &= \nabla_{\boldsymbol{\xi}_i}R_{i,j},
    &\tilde w_{i,j,g} &= \operatorname{Agg}_{\Gamma}(|\mathbf{g}_{i,j}|)_g,\\
    w_{i,j,g} &=
    \frac{\tilde w_{i,j,g}}
    {\max(\sum_{g'\in\mathcal{G}_\Gamma}\tilde w_{i,j,g'},\epsilon)},
    &g&\in\mathcal{G}_\Gamma.
\end{aligned}
\label{eq:credit_weights}
\end{equation}
Depending on $\Gamma$, $\operatorname{Agg}_\Gamma$ produces frame-level or spatially smoothed credit, falling back to uniform weights for invalid maps; granularity trades localization against noise (\suppref{app:token_credit_analysis}).

We call $|\mathbf{g}_{i,j}|$ the \emph{reward-sensitivity map} of check $j$: to first order, it measures how much the check's reward changes under a small perturbation of each video token; it is local and model-dependent, not a causal attribution. Unlike bottom-up saliency or foreground maps, it depends on the question and its reference answer rather than on the video alone: any score-independent map has zero question-swap interaction by construction, whereas ours is positive and consistent in sign (\suppref{app:question_position_diagnostic}). We reserve \emph{token credit} for the normalized routing weights $w_{i,j,g}$.

\subsection{Shared rollout advantage}
\label{sec:method_group_relative}

For each prompt, the old policy samples a group of $M$ rollouts. We first average the question rewards within each rollout and then normalize across the prompt group:
\begin{equation}
    \bar R_i=\frac{1}{K}\sum_{j=1}^{K}R_{i,j},
    \qquad
    A_i^{\mathrm{tvrl}}
    =\frac{\bar R_i-\mu_G}{\max(\sigma_G,\epsilon)},
    \label{eq:video_token_credit_advantage}
\end{equation}
where $\mu_G$ and $\sigma_G$ are the mean and population standard deviation over the $M$ rollouts. This order preserves the video-level ranking and avoids separate normalization statistics for each check. All checks share $A_i^{\mathrm{tvrl}}$ and route it through their own maps; a failed check can therefore receive a positive shared advantage. The maps allocate the video-level reward rather than supervise each question.

\subsection{Token-level policy update}
\label{sec:method_grpo}

Let $\Omega_g$ denote the entries of the dense transition log-probability map that belong to credit unit $g$. We reduce these entries, apply the question-specific credit map, and form a single weighted policy ratio:
\begin{equation}
\begin{aligned}
    u_{i,t,g}^{\theta}
    &=\frac{1}{|\Omega_g|}\sum_{r\in\Omega_g}
      \log\pi_\theta(\mathbf{z}_{i,t-1}^{r}\mid\mathbf{z}_{i,t},t,c),\\
    \ell_{i,t,j}^{\theta}
    &=\sum_{g\in\mathcal{G}_\Gamma}w_{i,j,g}u_{i,t,g}^{\theta},
    \qquad
    \rho_{i,t,j}(\theta)
    =\exp\!\left(\ell_{i,t,j}^{\theta}-\ell_{i,t,j}^{\theta_{\mathrm{old}}}\right).
\end{aligned}
\label{eq:weighted_ratio}
\end{equation}
Our token-level GRPO surrogate is
\begin{equation}
    \mathcal{L}_{\mathrm{tvrl}}
    =\frac{1}{TMK}\sum_{i,t,j}
    \max\!\left(
    -\hat A_i\rho_{i,t,j}(\theta),
    -\hat A_i\operatorname{clip}(\rho_{i,t,j}(\theta),1-\epsilon,1+\epsilon)
    \right),
    \label{eq:token_credit_loss}
\end{equation}
where $\hat A_i=\operatorname{clip}(A_i^{\mathrm{tvrl}},-A_{\max},A_{\max})$; we omit the inherited timestep-balancing factors for readability. Log-probability changes are aggregated before exponentiation and clipping. With uniform final latent-grid weights and identical check maps, the weighted log-probability reduces to the scalar baseline's reduction. By contrast, exponentiating and clipping a separate ratio for every unit would change the estimator, and this variant performs worse in our ablation (\suppref{app:additional_ablations}).

The weighted ratio is a local credit-assignment surrogate rather than the exact likelihood ratio of the full denoising trajectory. In the unclipped region, its gradient takes the form
\begin{equation}
    \nabla_\theta\mathcal{L}_{i,t,j}
    \propto
    -\hat A_i\rho_{i,t,j}(\theta)
    \sum_{g\in\mathcal{G}_\Gamma}
    w_{i,j,g}\nabla_\theta u_{i,t,g}^{\theta}.
    \label{eq:weighted_gradient_routing}
\end{equation}
Hence $\hat A_i$ determines the sign and overall magnitude of the update, while $w_{i,j,g}$ routes a larger share of it to the units most sensitive to check $j$. At the old policy every $\rho_{i,t,j}=1$, so to first order the $K$ maps act as their average $\bar w_{i,g}=\frac1K\sum_j w_{i,j,g}$; they differ only through exponentiation and clipping, and a single ratio with $\bar w$ performs worse (\suppref{app:additional_ablations}). \Suppref{app:token_credit_analysis} gives conditional intuitions for when routing helps.

\subsection{KL regularization}
\label{sec:method_kl}

As in SAGE-GRPO~\citep{zheng2026sagegrpo}, we add reference and stepwise old-policy KL terms:
\begin{equation}
\begin{aligned}
\mathcal D_{\mathrm{ref}}
&=\mathbb E_{i,t}\!\left[
\frac{\left\langle\|\boldsymbol\mu_\theta-\boldsymbol\mu_{\mathrm{ref}}\|_2^2\right\rangle}
{2\sigma_{\eta,t}^2}\right],\\
\mathcal D_{\mathrm{step}}
&=\mathbb E_{i,t}\!\left[
\log\pi_{\theta_{\mathrm{old}}}(\mathbf z_{i,t-1}\mid\mathbf z_{i,t},t,c)
-\log\pi_\theta(\mathbf z_{i,t-1}\mid\mathbf z_{i,t},t,c)\right],\\
\mathcal L
&=\mathcal L_{\mathrm{tvrl}}+\beta_{\mathrm{KL}}
\left(\lambda_{\mathrm{ref}}\mathcal D_{\mathrm{ref}}
+\lambda_{\mathrm{step}}\mathcal D_{\mathrm{step}}\right),
\end{aligned}
\label{eq:kl_regularization}
\end{equation}
where $\langle\cdot\rangle$ averages over non-batch entries and an adaptive controller updates $\beta_{\mathrm{KL}}$. Both regularizers remain global over transition states; token credit affects only the clipped policy-ratio term. The coefficients are listed in \suppref{app:training_settings}.
\endgroup

\begin{figure*}[t]
\centering
\includegraphics[width=\textwidth]{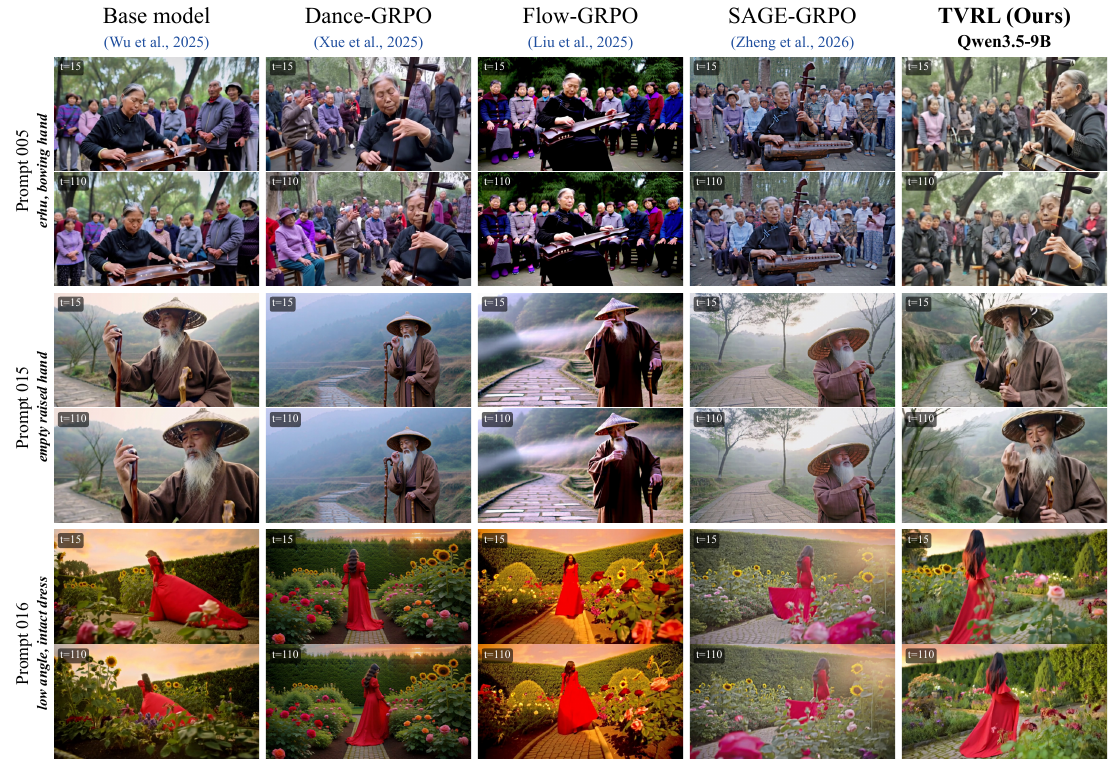}
\caption{\textbf{Qualitative results on HunyuanVideo-1.5.} Comparison of prompt- and seed-matched videos from the base model~\citep{hunyuan1.5}, Dance-GRPO~\citep{DanceGRPO}, Flow-GRPO~\citep{Flow-GRPO}, SAGE-GRPO~\citep{zheng2026sagegrpo}, and \methodname{} with the Qwen3.5-9B critic on SAGE-GRPO validation prompts (full text in \suppref{app:qualitative_prompts}); row labels name the prompt requirement to check. Only \methodname{} satisfies all three, whereas the baselines substitute a flat zither, put a fist or an object in the raised hand, or lose the low-angle framing and the dress.}
\label{fig:quality-comparison}
\end{figure*}

\section{Experiments}
\label{sec:experiments}

\subsection{Experimental Setup}
\label{sec:exp_setup}

\newsavebox{\tvrlmainbox}
\sbox{\tvrlmainbox}{\scriptsize\setlength{\tabcolsep}{2pt}%
\begin{tabular}{llrrrrrrr}
\toprule
 \textbf{Reward} & \textbf{Method} & \textbf{Overall} & $\boldsymbol{\Delta}$ & Creat. & Comm. & Ctrl. & Human & Phys. \\
\midrule
Base model & -- & 54.09 & -- & 41.40 & 62.75 & 30.26 & 88.94 & 47.11 \\
\midrule
VideoAlign & GRPO & 54.18 & -- & 41.44 & 61.14 & 30.77 & 90.06 & 47.49 \\
 \citep{VideoAlign} & \methodname{} & {55.54} & \textbf{+1.36} & 45.11 & 61.16 & 32.09 & 90.21 & 49.15 \\
\midrule
VideoScore2 & GRPO & 54.66 & -- & 42.23 & 64.89 & 30.29 & 91.52 & 44.35 \\
 \citep{VideoScore2} & \methodname{} & {55.99} & \textbf{+1.33} & 42.08 & 64.60 & 31.33 & 90.79 & 49.13 \\
 \midrule
 UnifiedReward2 & GRPO & 54.82 & -- & 42.90 & 62.14 & 30.25 & 88.90 & 50.90  \\
 \citep{wang2025unified} & \methodname{} & 56.67 & {\bf +1.85} & 45.36 & 64.55 & 31.29 & 89.87 & 52.28 \\
 \midrule
Qwen3.5-9B & GRPO & 54.54 & - & 41.68 & 64.88 & 31.57 & 88.85 & 45.74 \\
 \citep{qwen3.5} & \methodname{} & \textbf{57.69} & \textbf{+3.15} & 47.36 & 64.31 & 31.64 & 90.76 & 54.37 \\

\bottomrule

\end{tabular}}%
\begin{wraptable}{r}{\wd\tvrlmainbox}
\vspace{-8mm}
\begin{minipage}{\wd\tvrlmainbox}
\caption{\raggedright VBench-2.0 results with the SAGE sampler. Per reward model, GRPO uses uniform routing and \methodname{} gradient-based token routing; $\Delta$ is \methodname{}'s Overall gain.}
\label{tab:main-results}
\vspace{0.2\baselineskip}
\usebox{\tvrlmainbox}
\end{minipage}
\vspace{-4mm}
\end{wraptable}

\paragraph{Training setup.}
We fine-tune HunyuanVideo-1.5~\citep{hunyuan1.5} on the public SAGE-GRPO prompts~\citep{zheng2026sagegrpo}, keeping the VAE decoder, the text encoder, and the VLM critic frozen. Training rollouts contain 81 frames at $352\!\times\!624$ resolution with 20 denoising steps; evaluation videos contain 121 frames at $480\!\times\!864$ with 40 steps. We use the Muon optimizer with learning rate $10^{-5}$ and weight decay $0.01$ on 64 NVIDIA A100 80GB GPUs, with a per-device batch size of 2 and a group size of 4. The default Qwen3.5-9B critic scores five offline-decomposed yes/no checks by teacher-forced answer log-probability. Their average forms the shared group-relative advantage, and their credit maps route it in the policy ratio.

\paragraph{Baselines and variants.}
\Cref{tab:main-results} fixes the SAGE sampler and compares four reward models. GRPO uses each scalar reward with uniform routing, whereas \methodname{} additionally uses the video-frame-input gradient of the same reward for token routing. We also vary the SDE sampler and ablate credit granularity and the critic (more controls in \suppref{app:additional_experiments}).

\paragraph{Evaluation.}
We report the VBench-2.0 Overall score together with its creativity, common sense, controllability, human action, and physics dimensions~\citep{zheng2025vbench}. Our pairwise human study uses the first 200 prompts from VideoGen-Eval~\citep{yang2025videogen} and three distinct seeds per model and prompt, yielding 600 videos per model; eight annotators provide blind win, loss, or tie judgments of text--video alignment. \Suppref{app:human_evaluation} describes the full protocol. All evaluated checkpoints are taken after 100 optimizer steps, and unless stated otherwise, table entries and optimization curves average three random seeds and std showed in \suppref{app:seed_stats}.

\subsection{Main Results}
\label{sec:main_results}

Across the tested configurations, \methodname{} improves over matched GRPO baselines with several stochastic samplers and differentiable evaluators. It raises VBench-2.0 Overall with the SAGE, Flow, and Dance samplers and yields positive gains for all four evaluators tested under SAGE. These results establish empirical compatibility in the evaluated settings; they do not imply sampler- or evaluator-independent guarantees.

\begin{wrapfigure}{r}{0.4\columnwidth}
    \vspace{-2em}
    \centering
    \includegraphics[width=\linewidth]{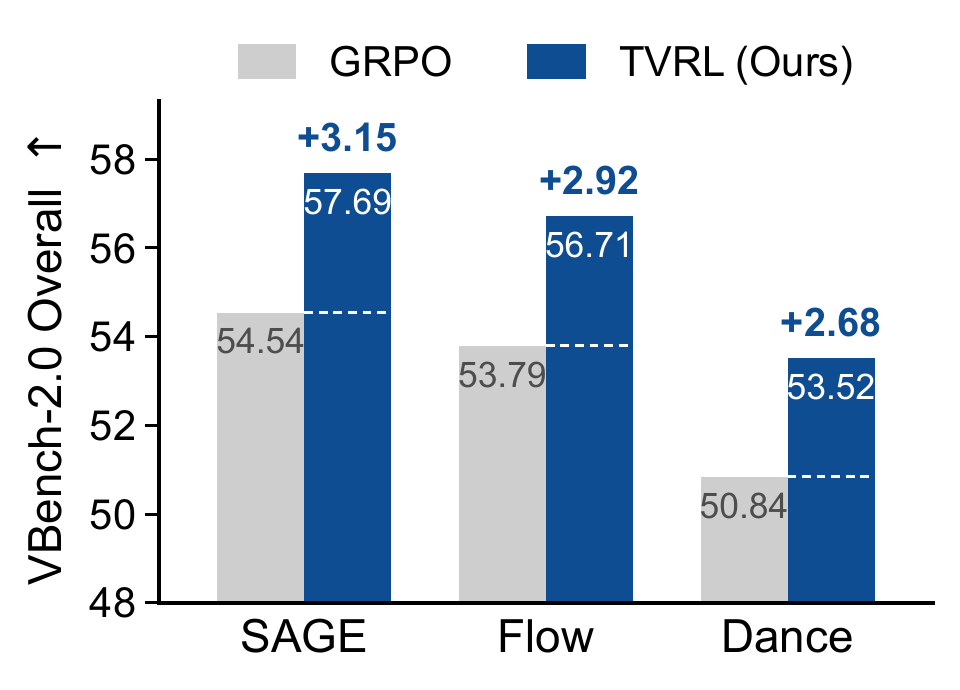}
    \vspace{-8mm}
    \caption{VBench-2.0 Overall of GRPO and \methodname{} across SDE samplers.}
    \label{fig:sde-comparison}
    \vspace{-9mm}
\end{wrapfigure}

\paragraph{SDE sampler compatibility.}

\Cref{fig:sde-comparison} fixes Qwen3.5-9B as the reward and credit model and varies the SDE sampler. \methodname{} raises Overall from 54.54 to 57.69 with SAGE (+3.15) and from 53.79 to 56.71 with Flow (+2.92). With Dance, the completed GRPO runs average 50.84, compared with 53.52 for \methodname{} (+2.68). Because the evaluator and the routing rule are held fixed, these gains indicate compatibility with different stochastic transitions rather than a change in the reward.

\paragraph{Reward model compatibility.}
\Cref{tab:main-results} holds the SAGE sampler fixed. \methodname{} improves Overall from 54.18 to 55.54 (+1.36) with VideoAlign, from 54.66 to 55.99 (+1.33) with VideoScore2, from 54.82 to 56.67 (+1.85) with UnifiedReward2, and from 54.54 to 57.69 (+3.15) with Qwen3.5-9B. Each pair uses the same evaluator and differs only in whether its video-frame-gradient map routes the policy update, so these three-seed means are consistent with a benefit from gradient-based routing.
With Qwen3.5-9B the gain comes mainly from creativity (+5.68) and physics (+8.63), while controllability is unchanged (31.57 to 31.64) and common sense drops slightly (64.88 to 64.31), suggesting that the size of the gain depends both on the scalar objective and on whether its video sensitivity provides useful routing.
More controls and results with other samplers are in \suppref{app:additional_experiments,app:videoalign_sde_results}.

\paragraph{Human preference.}
\Cref{tab:human-preference} summarizes the blind pairwise judgments. Participants prefer \methodname{} more often than the matched GRPO baseline under SAGE (36.3\% versus 23.3\%) and more often than the base model (31.3\% versus 19.2\%). The remaining 40.4\% and 49.5\% of judgments are ties, indicating that many pairs are perceptually comparable. Excluding ties, \methodname{} wins 60.9\% of the decisive comparisons against GRPO and 62.0\% against the base model, consistent with VBench-2.0.

\begin{wraptable}{r}{0.46\textwidth}
\centering
\vspace{-2em}
\caption{Blind pairwise human preferences for \methodname{} against the matched GRPO baseline under SAGE and the base model (\%).}
\label{tab:human-preference}
\small
\setlength{\tabcolsep}{4pt}
\begin{tabular}{lrrr}
\toprule
Pair & Win & Loss & Tie \\
\midrule
vs. GRPO (SAGE) & \textbf{36.3} & 23.3 & 40.4 \\
vs. Base & \textbf{31.3} & 19.2 & 49.5 \\
\bottomrule
\end{tabular}
\vspace{-2mm}
\end{wraptable}

\paragraph{Qualitative results.}
\Cref{fig:teaser-results} compares \methodname{} with the base model on prompt- and seed-matched examples: it renders a grand piano rather than a generic keyboard, places the requested denim jacket on a model, and corrects the twisted legs of a dancer. \Cref{fig:quality-comparison} compares against Dance-GRPO, Flow-GRPO, and SAGE-GRPO on three prompt- and seed-matched SAGE-GRPO validation prompts (full prompts in \suppref{app:qualitative_prompts}). The gains concentrate on the requirements the prompts spell out: only \methodname{} renders an upright erhu with the bow in motion, the man's empty hand counting on his fingers, and the low-angle garden shot with a coherent dress and the woman reaching toward a rose, whereas the baselines distort the erhu, lose the gesture, or collapse the dress. These examples illustrate, but do not substitute for, the quantitative and human studies.

\begin{figure*}[t]
\centering
\includegraphics[width=1.0\textwidth]{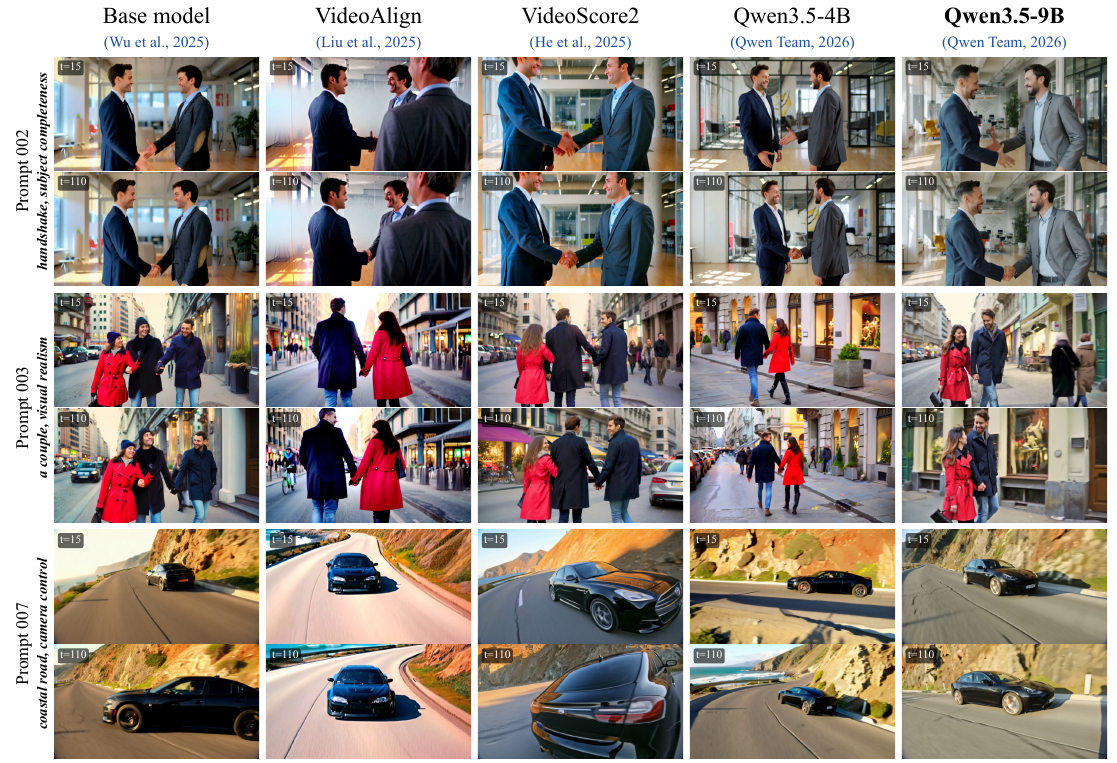}
\vspace{-2mm}
\caption{\textbf{Effect of the frozen critic.} Prompt- and seed-matched videos from the base model~\citep{hunyuan1.5} and \methodname{} with VideoAlign, VideoScore2, Qwen3.5-4B, or Qwen3.5-9B as the critic on VideoGen-Eval prompts~\citep{yang2025videogen} (\suppref{app:qualitative_prompts}); row labels name the aspect to check. Only the Qwen3.5-9B critic keeps all three, whereas weaker critics add a third person, show the couple from behind, or lose the road.}
\label{fig:critic-comparison}
\end{figure*}

\needspace{4\baselineskip}
\subsection{Ablation Study}
\label{sec:ablation}

\begin{wrapfigure}{r}{0.6\textwidth}
\vspace{-2em}
\centering
\includegraphics[width=0.495\linewidth]{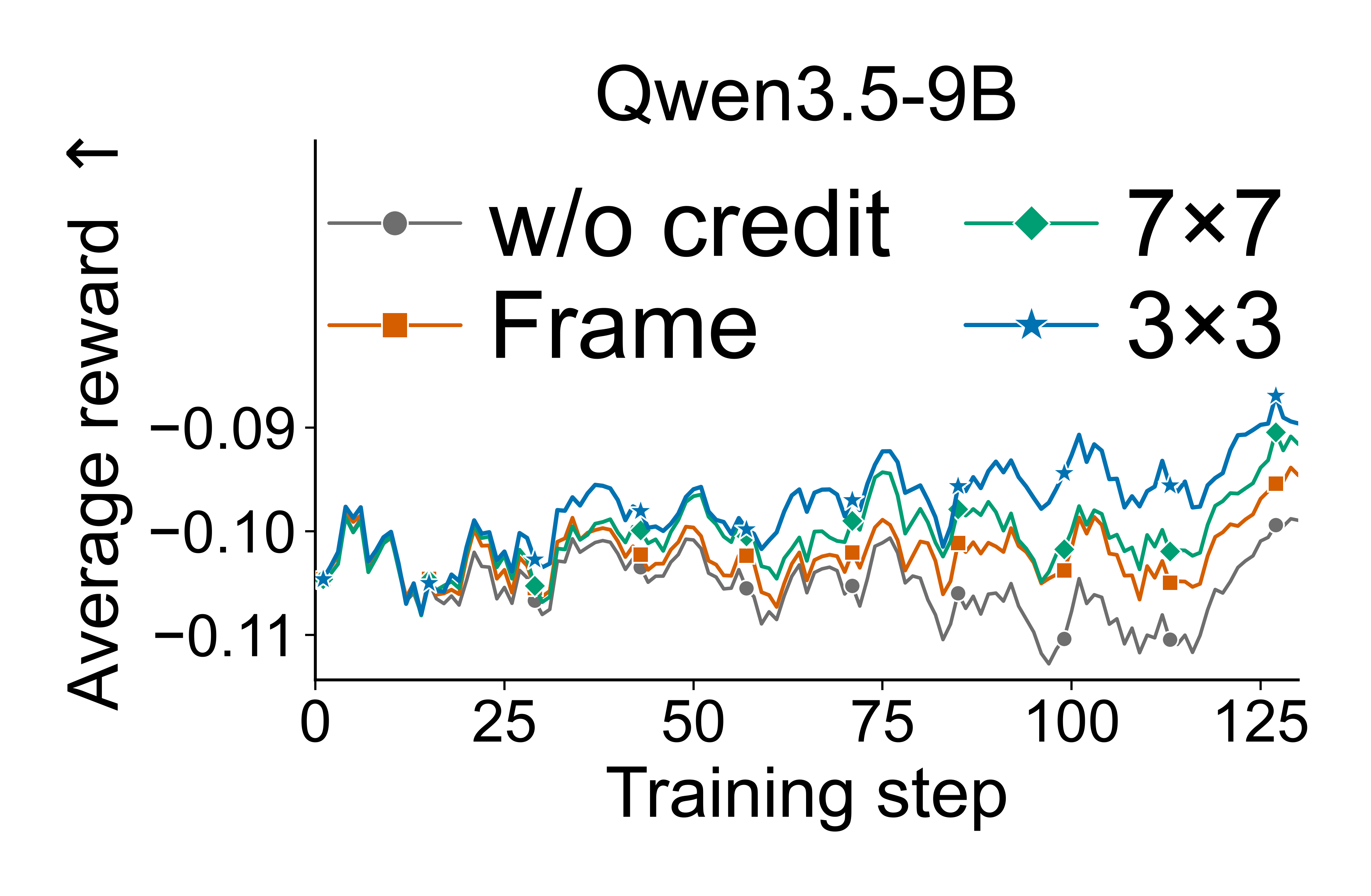}\hfill
\includegraphics[width=0.495\linewidth]{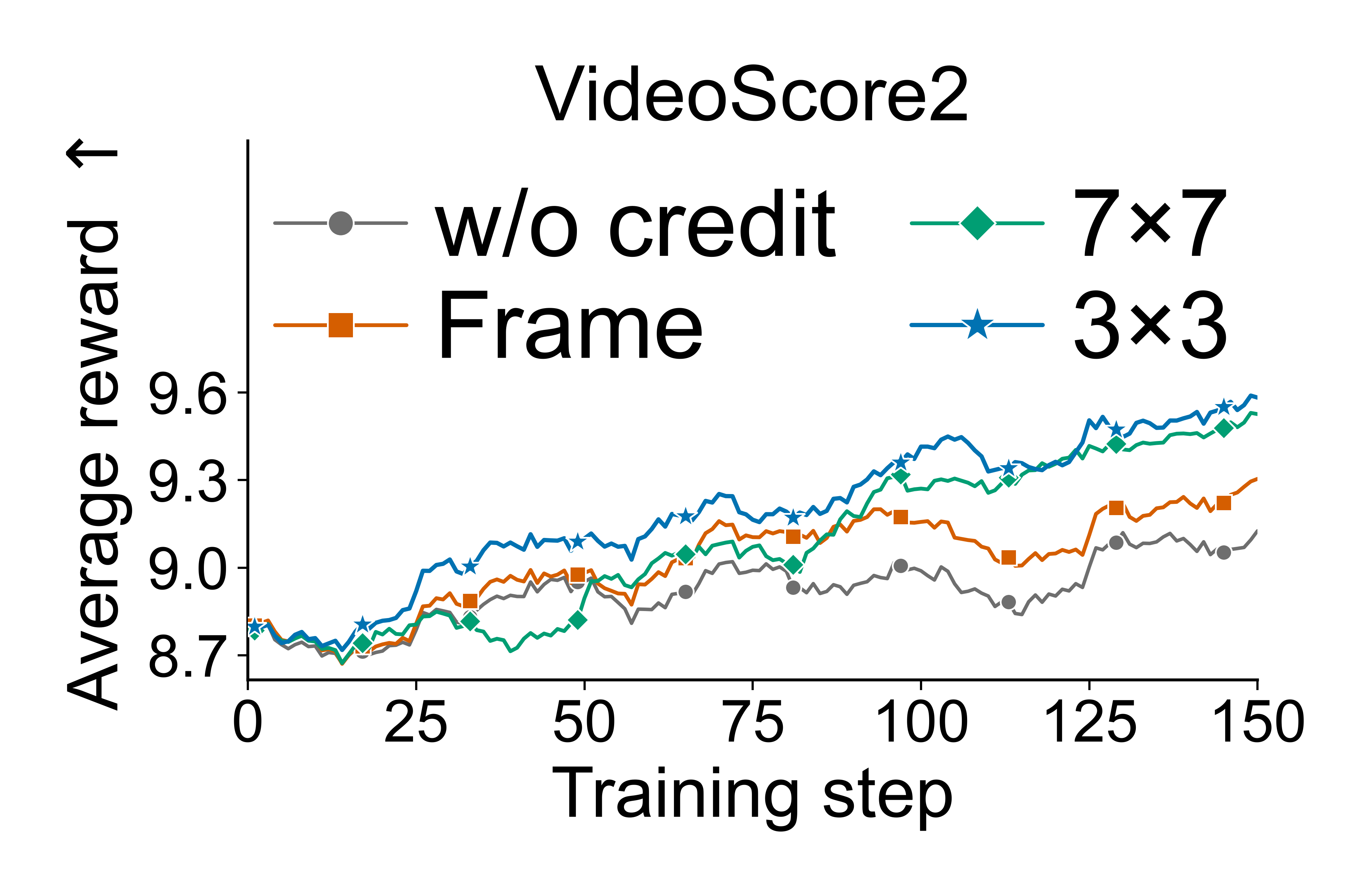}
\vspace{-6mm}
\caption{\textbf{Finer credit learns faster.} Training reward with Qwen3.5-9B (left) and VideoScore2 (right), averaged over three runs: w/o credit $<$ frame $<$ $7\!\times\!7$ $<$ $3\!\times\!3$ later in training.
}
\label{fig:reward-curves}
\end{wrapfigure}

\paragraph{Credit granularity.}
\Cref{fig:reward-curves} compares progressively localized credit: uniform weighting (w/o credit), frame-level credit, and $7\!\times\!7$ and $3\!\times\!3$ spatial credit. Averaged over three runs, the Qwen3.5-9B and VideoScore2 rewards both follow w/o credit $<$ frame $<$ $7\!\times\!7$ $<$ $3\!\times\!3$ in later training, which supports progressively localized routing within the tested range.
It does not imply that finer credit is always better: the unsmoothed $1\!\times\!1$ control reaches 55.10 Overall with Qwen3.5-9B, above uniform routing (54.54) but below frame-level credit (56.47; \suppref{app:cell_level_credit}).
\Cref{tab:reward-source-runs} provides complementary held-out evidence: Overall rises from 54.54 with uniform routing to 56.47 with frame credit, 57.08 with $7\!\times\!7$, and 57.69 with $3\!\times\!3$. Shuffling the $3\!\times\!3$ weights reduces Overall to 53.70, showing that the locations and coherence of the weights matter; since shuffling precedes smoothing and alters the final distribution, it does not isolate semantic alignment. We further use DINOv2 features as a reward-agnostic source of the routing map~\citep{ni2026seeingmattersvisualpreference}, keeping all other settings identical to our $3\!\times\!3$ setting. The result is 55.48 (DINOv2 map): nonuniform routing helps even without the reward, and deriving the map from the reward helps further.

\newsavebox{\tvrlroutingbox}\newsavebox{\tvrlcriticbox}

\sbox{\tvrlroutingbox}{\small\setlength{\tabcolsep}{3pt}%
        \vspace{-3em}
        \begin{tabular}{lrr}
            \toprule
            Routing & Overall & $\Delta$ \\
            \midrule
            Uniform & 54.54 & -- \\
            Frame & 56.47 & +1.93 \\
            $7\!\times\!7$ & 57.08 & +2.54 \\
            $3\!\times\!3$ shuffled & 53.70 & -0.84 \\
            DINOv2 map & 55.48 & +0.94 \\
            $3\!\times\!3$ \textbf{(Ours)} & \textbf{57.69} & \textbf{+3.15} \\
            \bottomrule
        \end{tabular}}%
\sbox{\tvrlcriticbox}{\small\setlength{\tabcolsep}{3pt}%
        \begin{tabular}{lrr}
            \toprule
            Critic & Overall & $\Delta$ \\
            \midrule
            Qwen3.5-0.8B & 51.85 & -5.84 \\
            InternVL3-1B & 53.98 & -3.71 \\
            Gemma4-E4B-it & 55.98 & -1.71 \\
            Qwen3.5-4B & 56.02 & -1.67 \\
            Qwen3.5-9B  & \textbf{57.69} & -- \\
            \bottomrule
        \end{tabular}}%
\begin{wraptable}{r}{\dimexpr\wd\tvrlroutingbox+\wd\tvrlcriticbox+1em\relax}
\vspace{-4.75mm}
\begin{minipage}[t]{\wd\tvrlroutingbox}
\setlength{\abovecaptionskip}{0pt}
        \localcaptionof{table}{\raggedright Routing ablation (SAGE, Qwen3.5-9B).}
        \label{tab:reward-source-runs}
        \vspace{0.2\baselineskip}
        \usebox{\tvrlroutingbox}
\end{minipage}\hfill
\begin{minipage}[t]{\wd\tvrlcriticbox}
\setlength{\abovecaptionskip}{0pt}
        \localcaptionof{table}{\raggedright Frozen VLM critic (SAGE).}
        \label{tab:backbone-model-scale-runs}
        \vspace{0.2\baselineskip}
        \usebox{\tvrlcriticbox}
\end{minipage}
\vspace{-3mm}
\end{wraptable}

\paragraph{Frozen VLM critic.}
\Cref{tab:backbone-model-scale-runs} tests sensitivity to the frozen VLM used jointly for reward and credit. Within the Qwen3.5 family, Overall rises monotonically from 51.85 with the 0.8B model to 56.02 with 4B and 57.69 with 9B. Critics from other families show a similar ordering: InternVL3-1B obtains 53.98 and Gemma4-E4B-it reaches 55.98. Because changing the critic changes both its scalar reward and its gradient map, this experiment cannot attribute the gain to either component alone; it shows that the complete \methodname{} signal benefits from the larger critics tested, not a general scaling law. \Cref{fig:critic-comparison} shows the same trend qualitatively.

\begin{wrapfigure}{r}{0.38\columnwidth}
\vspace{-2em}
\centering
\includegraphics[width=\linewidth]{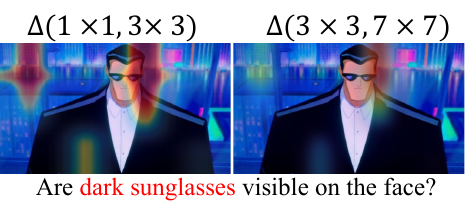}
\vspace{-3em}
\caption{Visualization of residuals $\Delta(1{\times}1,3{\times}3)$ and $\Delta(3{\times}3,7{\times}7)$.}
\label{fig:residual-credit}
\vspace{-1.2em}
\end{wrapfigure}

\paragraph{Spatial credit diagnostics.}
\Cref{fig:residual-credit} shows a qualitative example for \cref{tab:reward-source-runs} of why $3\!\times\!3$ credit is used by default. The $1\!\times\!1$ map is noisy: its residual over $3\!\times\!3$ (left) overemphasizes the background tokens unrelated to the check, consistent with the variance argument in \suppref{app:token_credit_analysis}. The $3\!\times\!3$ map is sharper than $7\!\times\!7$: its residual (right) lies near the key of the question, the sunglasses.

\section{Conclusion}

We presented \methodname{}, a framework that derives token-level credit from the reward being optimized in video reinforcement learning, and instantiated it in GRPO for text-to-video diffusion. Averaged teacher-forced question rewards set the sign and strength of each update through a shared group-relative advantage, and detached VLM gradients set its spatio-temporal allocation inside the clipped policy ratio. On VBench-2.0, \methodname{} improves over matched GRPO baselines for all four reward models under SAGE (+1.33 to +3.15 Overall) and for all three samplers with Qwen3.5-9B (+2.68 to +3.15), and $3\!\times\!3$ spatial credit outperforms both coarser and unsmoothed $1\!\times\!1$ credit. \methodname{} shows a promising direction for video RL: moving from trajectory-level to token-level updates, so that each rollout's advantage acts on the video tokens that determine its reward. We expect token-level credit to extend to longer videos and other evaluators.

\bibliography{iclr2027_conference}
\bibliographystyle{iclr2027_conference}

\clearpage
\appendix
\crefalias{section}{appendix}
\crefalias{subsection}{appendix}
\numberwithin{figure}{section}
\numberwithin{table}{section}
\numberwithin{equation}{section}
\renewcommand{\thefigure}{\thesection\arabic{figure}}
\renewcommand{\thetable}{\thesection\arabic{table}}
\renewcommand{\theequation}{\thesection\arabic{equation}}
\suppressfloats[t]
\newtcblisting{appendixpromptbox}[1][]{%
  listing only,
  breakable,
  colback=black!5!white,
  colframe=black!55!white,
  boxrule=0.4pt,
  arc=2pt,
  left=5pt,
  right=5pt,
  top=5pt,
  bottom=5pt,
  listing options={
    basicstyle=\ttfamily\footnotesize,
    columns=fullflexible,
    keepspaces=true,
    breaklines=true,
    breakatwhitespace=false,
    showstringspaces=false
  },
  #1
}
\section{Additional Experimental Results}
\label{app:additional_experiments}

\subsection{Human evaluation protocol}
\label{app:human_evaluation}

We use the first 200 prompts of VideoGen-Eval~\citep{yang2025videogen}. For every prompt, each compared model generates three videos from three distinct seeds, yielding 600 videos per model. The two pairwise comparisons evaluate \methodname{} against the matched GRPO baseline under SAGE and against the pretrained base model. Eight annotators complete the blind study with model identities hidden and judge text--video alignment as a win for \methodname{}, a loss, or a tie. We aggregate all judgments into the win/loss/tie rates in \mainref{tab:human-preference}; the decisive preference rate excludes ties and is computed as $\mathrm{Win}/(\mathrm{Win}+\mathrm{Loss})$.

\subsection{Prompts for the qualitative comparison}
\label{app:qualitative_prompts}

The three prompts in \mainref{fig:quality-comparison} are validation prompts 005, 015, and 016 of the public SAGE-GRPO prompt set~\citep{zheng2026sagegrpo}, and the three prompts in \mainref{fig:critic-comparison} are prompts 002, 003, and 007 of VideoGen-Eval~\citep{yang2025videogen}. Within each figure, all methods generate from the same prompt and seed. The full prompt text follows.

\paragraph{Prompt 005 (SAGE-GRPO validation).}
\begin{appendixpromptbox}
The scene opens with a medium shot of an elderly woman, appearing to be in her 80s, seated on a simple wooden stool in a public park. She has fine, silver hair neatly tied back and wears a traditional, dark-colored tunic with intricate embroidery at the collar. Her eyes are gently closed as she focuses on her music. In her hands, she holds a polished, dark wood erhu. Her left hand delicately presses the strings, her fingers moving with practiced grace, while her right hand draws the bow smoothly across them, producing a soulful, resonant melody. She hums softly, a low, gentle sound that harmonizes perfectly with the instrument's voice. A diverse crowd of onlookers, ranging from young adults to other seniors, has formed a respectful semi-circle around her. Some watch with rapt attention, their expressions contemplative. A young woman can be seen subtly recording on her phone, while an older man nods his head slowly to the rhythm. The park setting is serene, with large, mature trees whose leaves create a soft, dappled canopy. The light is diffused, characteristic of an overcast day, which lends a moody and introspective quality to the entire scene. The focus remains tightly on the woman, a figure of serene artistry amidst the quiet reverence of her audience.
\end{appendixpromptbox}

\paragraph{Prompt 015 (SAGE-GRPO validation).}
\begin{appendixpromptbox}
The scene opens on a sweeping wide shot, composed according to the rule of thirds, with an elderly Chinese man positioned on the right side of the frame. He stands on an ancient, weathered stone path that meanders through a landscape of rolling hills shrouded in a thick, soft mist. The man has a deeply wrinkled face and a long, flowing white beard that reaches his chest. He wears a simple, coarse-spun brown robe and a traditional conical bamboo hat that casts a gentle shadow over his features. In his left hand, he grips a gnarled wooden cane, its surface polished smooth from years of use, providing him with stability. His entire focus is on his right hand, which is held up near his face. His brow is furrowed in concentration as his thumb methodically touches different segments of his fingers in a complex, silent calculation. The environment is lit by a diffused, natural light coming from the side, which catches the airborne mist and softly illuminates the contours of his robe and the sparse, gnarled trees dotting the background. The atmosphere is one of profound peace and deep contemplation, as if the man is solving a timeless riddle alone in nature.
\end{appendixpromptbox}

\paragraph{Prompt 016 (SAGE-GRPO validation).}
\begin{appendixpromptbox}
The scene opens with a low-angle, full shot of a woman with dark, flowing hair, wearing a striking, floor-length scarlet red dress. She is strolling leisurely from right to left along a winding cobblestone path in a magnificent, verdant garden. The garden is in full bloom, with an array of colorful flowers like deep pink roses and tall, yellow sunflowers flanking the path. The camera is positioned low to the ground, making the flowers in the foreground appear large and immersive, framing the woman majestically. It's the golden hour; the setting sun is out of frame but casts long shadows and bathes the scene in a warm, golden light. This backlights the woman, creating a soft, ethereal glow around her silhouette and illuminating the delicate fabric of her dress. As she walks, she gently brushes her fingertips against the petals of a rose. The camera pans smoothly, following her elegant and unhurried pace. The background consists of tall green hedges and a soft-focus view of the sky, painted with hues of orange and pink.
\end{appendixpromptbox}

\paragraph{Prompt 002 (VideoGen-Eval).}
\begin{appendixpromptbox}
In a sleek, modern office with glass walls and minimalist decor, two men stand facing each other, their expressions radiating warmth and camaraderie. One man, wearing a tailored navy suit with a crisp white shirt, extends his hand with a genuine smile. The other, dressed in a charcoal blazer and light blue shirt, reciprocates the gesture, their hands meeting in a firm, enthusiastic handshake. Sunlight streams through large windows, casting a soft glow on the polished wooden floor and contemporary furniture. The atmosphere is one of mutual respect and shared success, captured in this moment of professional connection.
\end{appendixpromptbox}

\paragraph{Prompt 003 (VideoGen-Eval).}
\begin{appendixpromptbox}
A couple strolls hand in hand down a bustling city street, their fingers intertwined, embodying a sense of unity and affection. The man, wearing a navy coat and jeans, walks confidently beside the woman, who is dressed in a stylish red trench coat and black boots. The street is alive with the sounds of distant traffic and the chatter of passersby, while the couple remains in their own world, exchanging smiles and glances. The static camera captures their every step, as they navigate the urban landscape, framed by towering buildings and vibrant storefronts, under a sky painted with the warm hues of a setting sun.
\end{appendixpromptbox}

\paragraph{Prompt 007 (VideoGen-Eval).}
\begin{appendixpromptbox}
A sleek black car glides smoothly along a winding coastal road, the camera capturing its every move with a handheld tracking shot. The sun casts a golden glow on the vehicle's polished surface, highlighting its aerodynamic design. As the car navigates sharp turns, the ocean waves crash against the rocky cliffs below, creating a dramatic backdrop. The camera follows closely, capturing the rhythmic motion of the wheels and the subtle reflections of the surrounding landscape on the car's glossy exterior. The scene conveys a sense of freedom and adventure, with the open road stretching endlessly ahead.
\end{appendixpromptbox}

\subsection{Seed statistics}
\label{app:seed_stats}

\providecommand{\pmstd}[1]{\,{\scriptsize$\pm$#1}}
\Cref{tab:main-results-std,tab:ablation-std} report the mean $\pm$ standard deviation behind the main-text numbers, over $n$ runs that use independent training seeds and independent generation seeds, each evaluated at step 100 on the same prompt set.

\begin{table}[h]
\centering
\scriptsize
\setlength{\tabcolsep}{3pt}
\caption{\mainref{tab:main-results} with standard deviations over seeds (SAGE sampler, step 100).}
\label{tab:main-results-std}
\begin{tabular}{llcccccc c}
\toprule
Reward & Method & Overall & Creat. & Comm. & Ctrl. & Human & Phys. & $n$ \\
\midrule
Base model & -- 
& 54.09\pmstd{0.11} 
& 41.40\pmstd{0.38} 
& 62.75\pmstd{0.24} 
& 30.26\pmstd{0.46} 
& 88.94\pmstd{0.13} 
& 47.11\pmstd{0.34} 
& 3 \\
\midrule

\multirow{2}{*}{VideoAlign} 
& GRPO 
& 54.18\pmstd{0.23} 
& 41.44\pmstd{0.41} 
& 61.14\pmstd{0.26} 
& 30.77\pmstd{0.44} 
& 90.06\pmstd{0.12} 
& 47.49\pmstd{0.35} 
& 3 \\

& \methodname{} 
& 55.54\pmstd{0.42} 
& 45.11\pmstd{0.36} 
& 61.16\pmstd{0.27} 
& 32.09\pmstd{0.43} 
& 90.21\pmstd{0.14} 
& 49.15\pmstd{0.32} 
& 3 \\
\midrule

\multirow{2}{*}{VideoScore2} 
& GRPO 
& 54.66\pmstd{0.16} 
& 42.23\pmstd{0.39} 
& 64.89\pmstd{0.22} 
& 30.29\pmstd{0.48} 
& 91.52\pmstd{0.11} 
& 44.35\pmstd{0.37} 
& 3 \\

& \methodname{} 
& 55.99\pmstd{0.33} 
& 42.08\pmstd{0.40} 
& 64.60\pmstd{0.21} 
& 31.33\pmstd{0.45} 
& 90.79\pmstd{0.13} 
& 49.13\pmstd{0.31} 
& 3 \\
\midrule

\multirow{2}{*}{UnifiedReward2} 
& GRPO 
& 54.82\pmstd{0.47} 
& 42.90\pmstd{0.37} 
& 62.14\pmstd{0.25} 
& 30.25\pmstd{0.49} 
& 88.90\pmstd{0.15} 
& 50.90\pmstd{0.30} 
& 3 \\

& \methodname{} 
& 56.67\pmstd{0.19} 
& 45.36\pmstd{0.34} 
& 64.55\pmstd{0.23} 
& 31.29\pmstd{0.46} 
& 89.87\pmstd{0.12} 
& 52.28\pmstd{0.28} 
& 3 \\
\midrule

\multirow{2}{*}{Qwen3.5-9B} 
& GRPO 
& 54.54\pmstd{0.38} 
& 41.68\pmstd{0.42} 
& 64.88\pmstd{0.20} 
& 31.57\pmstd{0.44} 
& 88.85\pmstd{0.16} 
& 45.74\pmstd{0.36} 
& 3 \\

& \methodname{} 
& 57.69\pmstd{0.13} 
& 47.36\pmstd{0.33} 
& 64.31\pmstd{0.22} 
& 31.64\pmstd{0.47} 
& 90.76\pmstd{0.11} 
& 54.37\pmstd{0.27} 
& 3 \\
\bottomrule
\end{tabular}
\end{table}
\begin{table}[h]
\centering
\scriptsize
\setlength{\tabcolsep}{4pt}
\caption{Overall mean $\pm$ standard deviation for the sampler comparison (\mainref{fig:sde-comparison}), the routing ablation (\mainref{tab:reward-source-runs}), and the critic ablation (\mainref{tab:backbone-model-scale-runs}), all with the Qwen3.5-9B reward. Dance GRPO averages two runs; all other rows average three.}
\label{tab:ablation-std}
\begin{tabular}{llc}
\toprule
 & Setting & Overall \\
\midrule
\multirow{6}{*}{Sampler} & SAGE, GRPO & 54.54\pmstd{0.38} \\
 & SAGE, \methodname{} & 57.69\pmstd{0.13} \\
 & Flow, GRPO & 53.79\pmstd{0.31} \\
 & Flow, \methodname{} & 56.71\pmstd{0.27} \\
 & Dance, GRPO & 50.84\pmstd{0.42} \\
 & Dance, \methodname{} & 53.52\pmstd{0.34} \\
\midrule
\multirow{6}{*}{Routing} & Uniform & 54.54\pmstd{0.38} \\
 & Frame & 56.47\pmstd{0.29} \\
 & $7\!\times\!7$ & 57.08\pmstd{0.20} \\
 & $3\!\times\!3$ & 57.69\pmstd{0.13} \\
 & $3\!\times\!3$ shuffled & 53.70\pmstd{0.46} \\
 & DINOv2 map & 55.48\pmstd{0.32} \\
\midrule
\multirow{5}{*}{Critic} & Qwen3.5-0.8B & 51.85\pmstd{0.41} \\
 & InternVL3-1B & 53.98\pmstd{0.37} \\
 & Gemma4-E4B-it & 55.98\pmstd{0.28} \\
 & Qwen3.5-4B & 56.02\pmstd{0.24} \\
 & Qwen3.5-9B & 57.69\pmstd{0.13} \\
\bottomrule
\end{tabular}
\end{table}

\subsection{Additional ablations}
\label{app:additional_ablations}

\paragraph{Reward objective and token-level credit.}
\Cref{tab:pre-video-adv-ablation-runs} compares several SAGE/Qwen3.5-9B reward-objective variants that precede the final shared-advantage setting. Free-generation scoring, target log-probability scoring, and token credit obtain similar Overall scores, ranging from 54.28 to 54.54. The teacher-forced reward alone therefore does not explain the main-result gain. The stronger result in \mainref{tab:main-results} comes from combining VLM answer likelihoods with the shared rollout-level advantage and weighted-log-probability token credit.

\begin{table}[t]
\centering
\scriptsize
\setlength{\tabcolsep}{3pt}
\caption{Reward-objective ablations preceding the final shared-advantage setting. All rows use SAGE-GRPO with Qwen3.5-9B and the same decomposed binary-question prompts, while varying VLM scoring mode and temporal credit assignment.}
\label{tab:pre-video-adv-ablation-runs}
\resizebox{\linewidth}{!}{%
\begin{tabular}{lllrrrrrr}
\toprule
Variant & Reward & Credit & Overall & Creat. & Comm. & Ctrl. & Human & Phys. \\
\midrule
Free generation & \texttt{Free-gen} & Uniform & 54.28 & 37.32 & 67.75 & 29.75 & 87.42 & 49.14 \\
Log-probability reward & \texttt{Logprob} & Uniform & 54.54 & 41.68 & 64.88 & 31.57 & 88.85 & 45.74 \\
Token credit & \texttt{Logprob} & Per-question gradient & 54.42 & 39.98 & 66.31 & 30.38 & 88.41 & 47.02 \\
\bottomrule
\end{tabular}}
\end{table}

\paragraph{Reward-frame and question controls.}
\Cref{tab:weighting-prompt-frame-rating-runs} examines how much video and semantic evidence the critic uses. Increasing the number of sampled reward frames from 5 to 30 improves Overall from 53.49 to 55.71, with the largest gains in creativity, human action, and physics. Increasing the number of decomposed questions from 1 to 3 produces an even larger Overall gain, from 51.34 to 55.69, suggesting that broader semantic coverage matters for both reward estimation and credit assignment. The rating-threshold-5 variant reaches 55.41 Overall; because only one threshold is reported, we treat this as a sensitivity result rather than evidence for an optimal threshold.

\begin{table}[t]
\centering
\scriptsize
\setlength{\tabcolsep}{3pt}
\caption{Rating-threshold, reward-frame-count, and question-count ablations for SAGE-GRPO with Qwen3.5-9B token credit, evaluated after 100 optimization steps. All score columns report VBench-2.0 results.}
\label{tab:weighting-prompt-frame-rating-runs}
\resizebox{0.8\linewidth}{!}{%
\begin{tabular}{llrrrrrr}
\toprule
Control & Setting & Overall & Creat. & Comm. & Ctrl. & Human & Phys. \\
\midrule
Rating threshold & 5 & 55.41 & 36.98 & 69.20 & 33.03 & 88.71 & 49.14 \\
\midrule
\multirow{2}{*}{Reward frames} & 5  & 53.49 & 40.81 & 65.47 & 30.42 & 87.08 & 43.64 \\
& 30 & 55.71 & 44.60 & 61.37 & 30.41 & 90.02 & 52.14 \\
\midrule
\multirow{2}{*}{Questions} & 1 & 51.34 & 33.52 & 64.34 & 27.91 & 87.62 & 43.30 \\
& 3 & 55.69 & 40.24 & 65.43 & 31.58 & 90.20 & 51.00 \\
\bottomrule
\end{tabular}}
\end{table}

\paragraph{Gradient aggregation and ratio form.}
\Cref{tab:semantic-gradient-aggregation-runs} compares alternatives at frame-level credit, where per-question routing reaches 56.47. Because every ratio equals one at the old policy, per-question maps and their average $\bar w$ give the same first-order update; the averaged-map single ratio isolates what separate ratios add through exponentiation and clipping, and it drops to 54.96. A generic check in place of the prompt-specific ones also reaches 54.96, and clipping a separate ratio for each frame drops to 52.95, supporting a single weighted ratio per check (\mainref{sec:method_grpo}).

\begin{table}[t]
\centering
\scriptsize
\setlength{\tabcolsep}{3pt}
\caption{Ablations on how VLM-gradient credit is constructed and applied, for SAGE-GRPO with Qwen3.5-9B evaluated after 100 optimization steps. All rows use frame-level credit; per-question frame credit reaches 56.47. General semantic credit uses one generic check; averaged map uses one ratio with $\bar w=\frac1K\sum_j w_j$, the first-order equivalent of per-question ratios; per-frame ratio clips frame-wise ratios before aggregation.}
\label{tab:semantic-gradient-aggregation-runs}
\resizebox{0.8\linewidth}{!}{%
\begin{tabular}{lrrrrrr}
\toprule
Ablation & Overall & Creat. & Comm. & Ctrl. & Human & Phys. \\
\midrule
General semantic credit & 54.96 & 44.00 & 66.31 & 27.41 & 88.46 & 48.64 \\
Averaged map, single ratio & 54.96 & 42.40 & 64.88 & 34.71 & 86.94 & 45.88 \\
Per-frame ratio & 52.95 & 38.77 & 66.03 & 29.04 & 84.66 & 46.26 \\
\bottomrule
\end{tabular}}
\end{table}

\subsection{Training overhead}
\label{app:training_overhead}

\methodname{} incurs approximately 18\% higher observed time per update in our historical SAGE runs with VideoAlign (\cref{tab:training-overhead}); this pair does not measure the cost of the default Qwen3.5-9B configuration. We compare the scalar-reward GRPO baseline with the six-frame VideoAlign credit implementation using the shared rollout-level advantage and weighted-log-probability routing. Both runs use seed 42, batch size 2 per device, group size 4, sequence-parallel size 2, and 20 denoising steps, with the same logged optimizer and KL settings. We compute elapsed time from consecutive training-record timestamps over updates 11--100, excluding the first ten updates and retaining all 90 subsequent intervals. This measurement includes work outside the rollout and optimizer-loop timers, as well as any logging or checkpoint activity within these intervals.

\begin{table}[tbp]
\centering
\small
\caption{Observed training overhead in historical SAGE/VideoAlign runs. Times are seconds per update over updates 11--100; relative time is the ratio of means. Configuration differences are described in \cref{app:training_overhead}.}
\label{tab:training-overhead}
\begin{tabular}{lrrr}
\toprule
Method & Mean (s) & Median (s) & Relative time \\
\midrule
GRPO & 666.05 & 661.72 & $1.000\times$ \\
\methodname{} & 786.04 & 781.53 & $1.180\times$ \\
\bottomrule
\end{tabular}
\end{table}

The observed mean increases from 666.05 to 786.04 seconds per update, a difference of 119.99 seconds (18.02\%). This is an implementation-level wall-clock comparison, not an isolated measurement of VLM backward cost. In particular, the credit implementation also recomputes old-policy log-probabilities for weighted routing; this required work is included in the reported elapsed time. The comparison uses existing runs rather than a controlled timing rerun: the baseline and credit runs use VideoAlign MQ/TA/VQ reward weights of $(0.5,1,0.5)$ and $(1,1,1)$, respectively, and different prompt-file variants (original versus question-decomposed). The baseline reward-frame count is inherited from its reward-model defaults, while the credit run explicitly uses six frames; identical historical frame sampling, source revisions, and execution conditions have not been verified. We therefore report 18.02\% as the observed overhead for this pair, rather than a universal overhead estimate or evidence of a wall-clock speedup.

\subsection{Question-conditioned spatial sensitivity}
\label{app:question_position_diagnostic}

\paragraph{Relation to the training control.}
The shuffled $3\!\times\!3$ control in \mainref{tab:reward-source-runs} obtains 53.70 Overall, compared with 57.69 for unshuffled routing and 54.54 for uniform routing. In the local implementation, shuffling permutes routing-grid weights before spatial smoothing, preserving their value distribution at that stage; subsequent smoothing can change the distribution. This comparison supports structured routing but does not isolate semantic correspondence from spatial and temporal coherence or match the final maps' entropy. We separately test whether the critic's video sensitivity follows queried content when its position changes.

\paragraph{Controlled question and position swaps.}
We fix 12 disjoint pairs of source prompts and two generation seeds per pair, yielding 24 video pairs. Each source contributes 20 uniformly sampled frames, center-cropped to a square and resized to fill an equal-size panel; the crop is fixed and not object-aware. An earlier construction instead letterboxed each full frame into its panel with neutral padding, which left the content in roughly half of the panel area; we report it as a second row on the same pairs and questions. For each pair, we concatenate the panels as $AB$ and $BA$ and ask separately whether target $A$ or target $B$ is present. The questions do not specify positions. This gives 96 real Qwen3.5-9B gradient computations per panel construction of the teacher-forced log-probability of the fixed answer \texttt{Yes}. Unlike the full training question set, this diagnostic uses object-presence checks. Before computing gradients, we review three frames per pair and require each target to be visible in at least two. This qualifies 20 crop-to-fill pairs: one pair contains an ambiguous piano-like object, and in three pairs the fixed crop removes the target from at least two reviewed frames. The letterbox construction qualifies 23 pairs under the same rule. Unqualified pairs are retained in the all-pairs analysis, every prompt-pair group keeps at least one qualified pair, and no pairs are replaced after observing gradients. The construction uses known panel boundaries rather than detector or segmenter masks; each panel nevertheless contains background as well as the target.

\paragraph{Coordinate reconstruction and endpoint.}
The diagnostic uses the training critic configuration: 20 frames under a 50,176-pixel budget shared across the clip. The actual critic input is therefore $20\!\times\!32\!\times\!64$, producing a $10\!\times\!2\!\times\!4$ gradient grid in which each panel spans only $2\!\times\!2$ patches per temporal slot. We invert the processor's block-major patch packing to recover spatial coordinates, then apply the training helpers' interpolation to $20\!\times\!16\!\times\!16$ and $3\!\times\!3$ smoothing. Interpolation and smoothing leak credit across the panel boundary: a map placing all native-grid mass in one panel yields $D=84.8$ pp, not 100, so the endpoint is mildly attenuated but far from saturated. Let $L(q,AB)$ denote normalized credit mass in the left panel under question $q$ and layout $AB$. Define
\begin{equation}
 D=\tfrac12\bigl[L(A,AB)-L(B,AB)+L(B,BA)-L(A,BA)\bigr].
 \label{eq:question_position_interaction}
\end{equation}
Positive $D$ indicates that changing the question favors the corresponding panel, averaged over both positions. A score-independent saliency or foreground map gives zero within-layout differences; a fixed question-specific side preference cancels across layouts. Uniform routing has $D=0$, and mean queried-panel mass is $1/2+D/2$. We average seeds within each prompt-pair group and weight the 12 groups equally. Confidence intervals use 10,000 group bootstrap resamples; two-sided sign-flip tests operate on group means, not individual frames or gradient calls.

\begin{table}[htbp]
\centering
\caption{Question--position crossover diagnostic with Qwen3.5-9B at the training critic resolution. $D$ measures question-induced panel-credit change in percentage points (pp); the final column is mean credit mass in the queried panel, which equals $50\%$ under uniform routing. Every cohort contains 12 prompt-pair groups. Crop-to-fill is the primary panel construction; letterbox is the earlier construction on the same pairs and questions.}
\label{tab:question-position-diagnostic}
\small
\setlength{\tabcolsep}{5pt}
\begin{tabular}{llrrr}
\toprule
Panels & Cohort & Pairs & $D$ (pp), 95\% CI & Queried mass \\
\midrule
Crop-to-fill & Prequalified & 20 & $+5.39\;[2.83,7.77]$ & 52.70\% \\
Crop-to-fill & All pairs & 24 & $+6.41\;[4.46,8.43]$ & 53.20\% \\
Letterbox & Prequalified & 23 & $+3.79\;[1.66,5.99]$ & 51.90\% \\
Letterbox & All pairs & 24 & $+3.74\;[1.63,5.90]$ & 51.87\% \\
\bottomrule
\end{tabular}
\end{table}

\paragraph{Results and scope.}
The qualified-cohort interaction is $+5.39$ percentage points (pp), with a 95\% interval of $[2.83,7.77]$ and sign-flip $p=0.0034$. Mean queried-panel mass is 52.70\%, only 2.70 pp above uniform. Ten of twelve groups have positive interactions, and both layout-specific contrasts are positive in 15 of 20 pairs. Including all 24 pairs gives $+6.41$ pp $[4.46,8.43]$ with all twelve groups positive ($p=0.0005$), and the letterbox construction gives a smaller effect of the same sign, $+3.79$ pp $[1.66,5.99]$ (\cref{tab:question-position-diagnostic}). The interaction is thus modest in magnitude but consistent in sign across groups and across both panel constructions, which supports question-conditioned regional sensitivity in this constructed setting; a score-independent map would give zero.

An exploratory decomposition, computed after the primary outcome was observed, indicates where the question dependence lies at this resolution. Within a fixed video, the two question-conditioned routed maps differ by a total variation of 30.5 pp on average. Their temporal marginals alone differ by 28.8 pp, whereas their spatial marginals differ by 8.3 pp, of which $D$ is the component directed at the queried panel. With a $2\!\times\!4$ spatial grid per temporal slot, changing the question therefore mainly reallocates credit across frames and only secondarily across space. This agrees with the training ablation in \mainref{tab:reward-source-runs}, where frame credit accounts for most of the gain over uniform routing (54.54 to 56.47 Overall) and $3\!\times\!3$ spatial credit adds a smaller increment (to 57.69).

Because each panel contains background as well as the target, the result does not establish precise object segmentation or causal credit for generator updates. The deliberately selected split-screen examples and the low critic resolution limit generalization; the bootstrap assumes exchangeable groups and the sign-flip test assumes sign symmetry under the null. This diagnostic neither compares gradients with attention nor establishes that $3\!\times\!3$ is superior to $7\!\times\!7$.

\section{Additional Implementation Details}
\label{app:implementation}

\subsection{Diffusion policy details}
\label{app:diffusion_policy_details}

Let $\mathbf{z}_T\sim\mathcal{N}(0,I)$. For a flow-matching velocity model, the deterministic Euler mean is
\begin{equation}
    \boldsymbol{\mu}_\theta(\mathbf{z}_t,t,c)
    =\mathbf{z}_t+\Delta t_t\,\mathbf{v}_\theta(\mathbf{z}_t,t,c).
\end{equation}
During GRPO rollouts, the scheduler wraps this transition in a stochastic SDE step,
\begin{equation}
    \mathbf{z}_{t-1}\sim
    \mathcal{N}\!\left(
    \boldsymbol{\mu}^{\mathrm{sde}}_\theta(\mathbf{z}_t,t,c),
    \sigma_{\eta,t}^{2}I
    \right),
    \label{eq:sde_transition}
\end{equation}
where $\eta$ controls the injected sampling noise. Up to terms independent of $\theta$, its transition log-probability is
\begin{equation}
    \log\pi_\theta(\mathbf{z}_{t-1}\mid\mathbf{z}_t,t,c)
    =-
    \frac{\|\mathbf{z}_{t-1}-\boldsymbol{\mu}^{\mathrm{sde}}_\theta(\mathbf{z}_t,t,c)\|_2^2}
    {2\sigma_{\eta,t}^{2}}
    +\mathrm{const.}
    \label{eq:diffusion_logprob}
\end{equation}
Different SDE samplers change the transition mean and the noise schedule but not the token-credit construction. We retain the unreduced elementwise log-probability map and aggregate it over the credit units as in \mainref{eq:weighted_ratio}. The KL-regularized training objective is defined in \mainref{eq:kl_regularization}.

\subsection{Training and reward settings}
\label{app:training_settings}

\Cref{tab:appendix-main-setting} summarizes the main \methodname{} setting used for the SAGE/Qwen3.5-9B token-credit video-adv run. Machine-specific paths, logging credentials, and storage locations are omitted. The Flow and Dance variants keep the same token-credit reward and GRPO configuration and change only the stochastic denoising sampler.

\begin{table}[h]
\centering
\small
\caption{Main training and reward settings.}
\label{tab:appendix-main-setting}
\resizebox{\linewidth}{!}{%
\begin{tabular}{ll}
\toprule
Component & Setting \\
\midrule
Generator & Text-to-video diffusion transformer, pretrained \texttt{480p\_t2v} checkpoint \\
Post-training objective & GRPO with \methodname{} token credit \\
Main SDE type & \texttt{sage\_grpo} \\
Reward critic & Frozen Qwen3.5-9B VLM \\
Reward model family & \texttt{qwen3\_5} \\
Reward score type & \texttt{token\_credit} \\
Reward VLM resolution budget & 50,176 max pixels \\
Reward VLM batch size & 1 generated video \\
Reward VLM generation cap & 128 new tokens \\
Prompt checks & Five binary yes/no video-grounded checks per prompt \\
Reward frames & 20 uniformly sampled frames per generated video \\
Group size & 4 rollouts per prompt \\
Training rollout & 81 frames at $352\!\times\!624$, 20 denoising steps \\
Evaluation generation & 121 frames at $480\!\times\!864$, 40 denoising steps \\
Hardware & 64 NVIDIA A100 80GB GPUs \\
Per-device batch size & 2 \\
Learning rate & $1\times 10^{-5}$ \\
Optimizer & Muon optimizer, weight decay $0.01$ \\
LR schedule & Constant \\
Training budget & 100 optimizer steps for every evaluated checkpoint \\
Timestep shift & 5.0 \\
Timestep permutation & Enabled \\
Gradient balancing & Enabled \\
Sequence parallel size & 2 \\
Memory options & FSDP and gradient checkpointing enabled \\
Loss mode & \texttt{video\_adv} \\
Ratio mode & \texttt{weighted\_logprob} \\
KL regularization & Moving reference KL and stepwise dual KL enabled \\
KL coefficients & KL weight $10^{-5}$, initial adaptive coefficient $10^{-7}$ \\
Dual KL weights & Reference-moving weight 1.0, old-policy step weight 0.1 \\
Reference update interval & 10 optimizer steps \\
Checkpoint interval & 25 optimizer steps in the current launcher \\
\bottomrule
\end{tabular}}
\end{table}

The training prompt file contains 10,000 text prompts. Each prompt is decomposed offline into an \texttt{overall\_question}, five \texttt{sub\_questions}, one structured VLM question prompt, and a structured reference answer. We do not use supervised target videos for post-training. During rollout, the generator samples videos for each text prompt, the frozen VLM scores the generated videos by teacher-forced answer likelihood, and the gradients of the same likelihoods provide the token-level credit weights.

\subsection{Evaluation protocol}
\label{app:evaluation_protocol}

All runs are evaluated with the VBench-2.0 automatic metrics on the same held-out prompt split for matched baselines and \methodname{} variants. Tables list Overall first, followed by creativity, common sense, controllability, human action, and physics.

\Needspace*{14\baselineskip}
\subsection{Supplementary VideoAlign sampler results}
\label{app:videoalign_sde_results}

\begin{wraptable}{r}{0.42\textwidth}
\centering
\vspace{-1.6em}
\caption{VBench-2.0 Overall with VideoAlign as the reward model across SDE samplers; $\Delta$ is the gain of \methodname{} over GRPO.}
\label{tab:videoalign-sde-supplement}
\vspace{0.2\baselineskip}
\small
\setlength{\tabcolsep}{5pt}
\begin{tabular}{llrr}
\toprule
SDE & Method & Overall & $\Delta$ \\
\midrule
\multirow{2}{*}{SAGE} & GRPO & 54.18 & -- \\
 & \methodname{} & 55.54 & +1.36 \\
\midrule
\multirow{2}{*}{Flow} & GRPO & 55.07 & -- \\
 & \methodname{} & 55.82 & +0.75 \\
\midrule
\multirow{2}{*}{Dance} & GRPO & 48.02 & -- \\
 & \methodname{} & 50.30 & +2.28 \\
\bottomrule
\end{tabular}
\vspace{-1em}
\end{wraptable}

\Cref{tab:videoalign-sde-supplement} repeats the sampler comparison of \mainref{fig:sde-comparison} with VideoAlign in place of Qwen3.5-9B as the reward model. \methodname{} improves over the matched GRPO run under all three samplers: from 54.18 to 55.54 with SAGE (+1.36), from 55.07 to 55.82 with Flow (+0.75), and from 48.02 to 50.30 with Dance (+2.28). The gains are smaller than with Qwen3.5-9B (+3.15, +2.92, and +2.68), consistent with the main-text observation that the size of the improvement depends on the reward model whose gradient supplies the routing signal.

\subsection{Prompt decomposition template}
\label{app:decomposition_prompt}

We use an offline language model to convert each text-to-video prompt into five atomic yes/no checks. The decomposition model receives only the text prompt, not a generated video. The system message is:

\begin{appendixpromptbox}[title={System message}]
You are a structured data generator for text-to-video evaluation. Return exactly one valid JSON
object and nothing else.
\end{appendixpromptbox}

The user template is:

\begin{appendixpromptbox}[title={Prompt decomposition template}]
You are an expert prompt-to-subquestion decomposition engine for text-to-video generation.

You are given a single text-to-video prompt, not a video and not an original evaluation question.
Your job is to convert the prompt itself into a compact binary checklist that can later be used to
verify whether a generated video faithfully follows the prompt.

Goal:
- Produce exactly <|NUM_SUBQUESTIONS|> atomic yes/no sub-questions.
- Each sub-question must target one concrete claim that is directly verifiable from the generated video and explicitly stated in
  the prompt.
- The sub-questions should together cover the main subject, scene, action or motion, key objects or
  attributes, and camera or composition when present.

Instructions:
1. Read the prompt carefully.
2. Rewrite the prompt intent as one concise `overall_question`.
3. Produce exactly <|NUM_SUBQUESTIONS|> sub-questions.
4. Every sub-question must be:
   - directly verifiable from a generated video,
   - directly grounded in the prompt text,
   - phrased as a clear yes/no question,
   - atomic and non-redundant.
5. Prefer concrete video-observable facts:
   - subject identity or category,
   - visible objects,
   - explicit colors or materials,
   - spatial relations,
   - explicit actions or motion,
   - explicit camera motion or framing,
   - visible environment or background.
6. Do not ask about hidden intent, symbolism, or facts that cannot be observed in the video.
7. Avoid subjective or weakly grounded wording such as:
   - beautiful
   - impressive
   - dramatic
   - immersive
   - majestic
   - lonely
   - surreal
   - dreamlike
   unless the prompt gives an explicit manifestation that can be checked directly in the video.
8. Do not ask multiple things in one sub-question.
9. Do not repeat the same fact using slightly different wording.
10. If the prompt contains camera language, at least one sub-question should cover it.
11. If the prompt does not contain camera language, replace that slot with another explicit video-observable
    fact from the prompt.
12. Every `answer_type` must be `yes_no`.
13. Do not output answers.
14. Never return fewer than <|NUM_SUBQUESTIONS|> sub-questions.
15. If a prompt clause is broad, split it into smaller atomic video-grounded checks so that the final output
    still contains exactly <|NUM_SUBQUESTIONS|> items.
16. Prefer distributing the <|NUM_SUBQUESTIONS|> slots across subject, appearance, action,
    environment, and camera/composition when those aspects are present.

Output format:
- Return valid JSON only.
- No markdown.
- No code fences.
- No explanation outside the JSON.

Return JSON with this exact schema:
{
  "overall_question": "...",
  "sub_questions": [
    {
      "id": "q1",
      "question": "...?",
      "answer_type": "yes_no"
    }
  ]
}

Hard constraints:
- `sub_questions` must contain exactly <|NUM_SUBQUESTIONS|> items.
- IDs must be `q1` to `q<|NUM_SUBQUESTIONS|>` in order.
- Every `question` must end with a question mark.
- Every `question` must be answerable by `Yes` or `No`.
- Every `answer_type` must be exactly `yes_no`.
- Returning 1, 2, 3, 4, or more than <|NUM_SUBQUESTIONS|> sub-questions is invalid.

Example input prompt:
A red fox leaps across a mossy fallen log in a misty forest while the camera tracks smoothly from
left to right.

Example output:
{
  "overall_question": "Does the generated video faithfully depict the main scene elements and
  motion described in the prompt?",
  "sub_questions": [
    {"id": "q1", "question": "Is a red fox visible in the video?", "answer_type": "yes_no"},
    {"id": "q2", "question": "Is the fox moving in a leaping or jumping motion?",
     "answer_type": "yes_no"},
    {"id": "q3", "question": "Is a mossy fallen log visible in the scene?",
     "answer_type": "yes_no"},
    {"id": "q4", "question": "Is the environment a misty forest?", "answer_type": "yes_no"},
    {"id": "q5", "question": "Does the camera track smoothly from left to right?",
     "answer_type": "yes_no"}
  ]
}

Now analyze this prompt:

<|PROMPT|>
\end{appendixpromptbox}

The decomposition is generated deterministically with a Qwen3.5-9B language model using temperature $0$ and top-$p$ of $1.0$. We request exactly five sub-questions for the main experiments. If the output is invalid JSON or violates the schema, the script retries with a repair prompt that repeats the original prompt, the invalid output, and the validation error.

\subsection{VLM reward prompt and reference answer}
\label{app:vlm_reward_prompt}

After decomposition, the sub-questions are formatted into a structured video QA prompt. This is the prompt text given to the VLM together with sampled video frames:

\begin{appendixpromptbox}[title={VLM reward prompt}]
You are a careful video question-answering assistant.
Use only evidence grounded in the video, and do not guess unseen details.

Given an AI-generated video, answer the following atomic video-grounded questions using only evidence from
the video.

Rules:
- Answer each question independently.
- Use concise canonical phrases.
- Do not add explanations.
- For "yes_no" questions, answer with "Yes" or "No" only.
- Return valid JSON only.

Output format:
{"sub_answers": [{"id": "q1", "answer": "..."}]}

Sub-questions:
- {"id": "q1", "question": "<question 1>", "answer_type": "yes_no"}
- {"id": "q2", "question": "<question 2>", "answer_type": "yes_no"}
- {"id": "q3", "question": "<question 3>", "answer_type": "yes_no"}
- {"id": "q4", "question": "<question 4>", "answer_type": "yes_no"}
- {"id": "q5", "question": "<question 5>", "answer_type": "yes_no"}
\end{appendixpromptbox}

The reference answer is appended as the assistant message and is used only for teacher-forced scoring:

\begin{appendixpromptbox}[title={Teacher-forced reference answer}]
{"sub_answers": [
  {"id": "q1", "answer": "Yes"},
  {"id": "q2", "answer": "Yes"},
  {"id": "q3", "answer": "Yes"},
  {"id": "q4", "answer": "Yes"},
  {"id": "q5", "answer": "Yes"}
]}
\end{appendixpromptbox}

The reward for a slot is the mean target-answer log-probability over the answer-value tokens in the reference answer. Slot rewards are averaged into the rollout reward before group-relative normalization. Credit weights come from the absolute gradient of each slot reward with respect to the processed VLM video-frame input, normalized across time and space for spatial routing (across time for frame-only routing). The maps remain question-conditioned but share one rollout-level advantage. All slots are evaluated in one teacher-forced sequence; later slots are conditioned on preceding reference tokens, not statistically independent answers. Answer-token predictions from that same forward pass are used only for monitoring: they neither select the fixed reference target nor enter the main reward. Autoregressive free-generation scoring is a separate ablation.

\subsection{Mathematical interpretation of token-credit routing}
\label{app:token_credit_analysis}

This section gives conditional intuitions, not guarantees, for the weighted-log-probability update. \mainref{eq:weighted_ratio} is not the exact likelihood ratio of the denoising trajectory, and each claim below holds only under the stated assumptions.

Consider one rollout $i$, one denoising step $t$, and one video-grounded check $j$. Ignoring clipping, the weighted-log-probability loss term is
\begin{equation}
    \mathcal{L}_{i,t,j}^{\mathrm{unclip}}
    =
    -\hat{A}_{i}
    \rho_{i,t,j}(\theta),
    \qquad
    \rho_{i,t,j}(\theta)
    =
    \exp\left(
    \sum_f w_{i,j,f}
    \left(
    u_{i,t,f}^{\theta}
    -
    u_{i,t,f}^{\theta_{\mathrm{old}}}
    \right)
    \right).
    \label{eq:app_unclipped_tc_loss}
\end{equation}
Because the old-policy log-probabilities and the frame weights are detached from the generator update, the gradient is
\begin{equation}
    \nabla_\theta \mathcal{L}_{i,t,j}^{\mathrm{unclip}}
    =
    -\hat{A}_{i}
    \rho_{i,t,j}(\theta)
    \sum_f
    w_{i,j,f}
    \nabla_\theta u_{i,t,f}^{\theta}.
    \label{eq:app_weighted_gradient}
\end{equation}
Thus the contribution of frame $f$ to the policy-gradient update routed through check $j$ is proportional to $w_{i,j,f}$. The shared advantage $\hat{A}_i$, computed from the average reward across checks, determines the sign and overall magnitude of the update: positive-advantage samples are reinforced and negative-advantage samples are suppressed, with the largest change assigned to high-credit frames. The map is a detached routing distribution, not an independent question-level advantage.

The first-order policy-improvement view explains when this routing is useful. For a small update from the old policy, the expected reward improvement can be approximated by
\begin{equation}
    \Delta J^{(1)}
    \approx
    \mathbb{E}
    \left[
    A(x)
    \Delta \log \pi(x)
    \right].
    \label{eq:app_first_order_improvement}
\end{equation}
If the video log-probability decomposes into frame-level changes $\Delta\log\pi_f$, standard scalar GRPO uses the full or uniformly averaged trajectory change, while \methodname{} uses the credit-weighted local surrogate
\begin{equation}
    \Delta \log \pi_{\mathrm{tvrl}}(x)
    =
    \sum_{f=1}^{F} w_f \Delta\log\pi_f.
    \label{eq:app_weighted_logprob_surrogate}
\end{equation}
This surrogate is biased relative to the exact trajectory likelihood, but it is beneficial when the weight vector is aligned with the frames that can actually change the reward.

To make this condition explicit, let $c_f\ge 0$ denote the true local reward sensitivity of frame $f$, and let $u_f$ denote the update magnitude assigned to that frame. Under a fixed KL-like update budget $\sum_f u_f^2\le B$, a frame-level linear model gives
\begin{equation}
    \Delta J_{\mathrm{frame}}^{(1)}
    \approx
    \sum_{f=1}^{F} c_f u_f
    \le
    \sqrt{B}\,\|c\|_2,
    \label{eq:app_budget_bound}
\end{equation}
where equality is achieved when $u_f\propto c_f$. By contrast, a uniform allocation gives
\begin{equation}
    \Delta J_{\mathrm{uniform}}^{(1)}
    =
    \sqrt{\frac{B}{F}}
    \sum_{f=1}^{F} c_f
    \le
    \sqrt{B}\,\|c\|_2,
    \label{eq:app_uniform_bound}
\end{equation}
with equality only when the reward sensitivity is uniform over frames. This is an idealized allocation problem: it identifies $u_f\propto c_f$ as optimal, but it does not show that unsigned VLM sensitivity induces such updates through a shared-parameter generator, nor that positive alignment alone beats uniform allocation.

For variance, compare against the uniform average $g_{\mathrm{unif}}=\frac{A}{F}\sum_f h_f$ with $h_f=\nabla_\theta\log\pi_\theta(x_f)$, which has the same scale as
\begin{equation}
    g_{\mathrm{tvrl}}
    =
    A
    \sum_{f=1}^{F}
    w_f
    \nabla_\theta \log \pi_\theta(x_f)
    \label{eq:app_weighted_estimator}
\end{equation}
For independent $h_f$ with $\operatorname{tr}\operatorname{Cov}(h_f)=\sigma_f^2$, weighting lowers variance only if $\sum_f w_f^2\sigma_f^2<F^{-2}\sum_f\sigma_f^2$. With equal $\sigma_f$ it cannot, since $\sum_f w_f^2\ge 1/F$; a reduction requires that low-credit frames carry more reward-irrelevant noise than high-credit ones.

The same variance argument also explains why arbitrarily fine credit need not be optimal. Let the estimated dense credit be $\hat{c}_g=c_g+\varepsilon_g$, where $c_g$ is the reward-relevant local sensitivity and $\varepsilon_g$ is model- and sample-dependent estimation noise. A small aggregation window introduces little spatial bias but retains more of $\varepsilon_g$; after nonnegative normalization, isolated responses can produce a concentrated weight vector. For weakly correlated unit-level policy gradients $h_g=\nabla_\theta\log\pi_\theta(x_g)$, the conditional covariance approximately satisfies
\begin{equation}
    \operatorname{tr}\operatorname{Cov}\!\left(A\sum_g w_g h_g\mid w\right)
    \approx
    A^2\sum_g w_g^2\operatorname{tr}\operatorname{Cov}(h_g).
    \label{eq:app_credit_granularity_variance}
\end{equation}
When the unit-level variances are comparable, a peaky $1\!\times\!1$ map has a smaller effective number of credited units, $N_{\mathrm{eff}}(w)=1/\sum_g w_g^2$, and can yield a higher-variance update. Increasing the spatial window averages local estimation noise and typically raises $N_{\mathrm{eff}}$, but an overly large window introduces localization bias by mixing distinct semantic regions. The preferred window therefore balances estimator variance against spatial bias; the analysis predicts an intermediate optimum rather than monotonic improvement as the credit map becomes finer.

Finally, PPO clipping does not change the meaning of the weights. Inside the unclipped region, \cref{eq:app_weighted_gradient} gives the exact local gradient of the surrogate. When clipping is active, the gradient from that sample--check pair is bounded or removed to avoid an overly large policy update, but the frame weights still specify where the active policy-gradient signal is routed.

This analysis is conditional: it does not establish that a VLM gradient is a perfect causal explanation of video semantics. The gradient is a local, model-dependent sensitivity measure, and its absolute value captures importance rather than sign. The practical claim is that when this sensitivity aligns with prompt-relevant temporal evidence, token credit concentrates the GRPO update on the frames where changes most likely affect the reward.

\suppressfloats[t]
\subsection{Unsmoothed cell-level credit}
\label{app:cell_level_credit}

\begin{figure*}[!htb]
\centering
\begin{minipage}{0.49\textwidth}
\centering
\includegraphics[width=\linewidth]{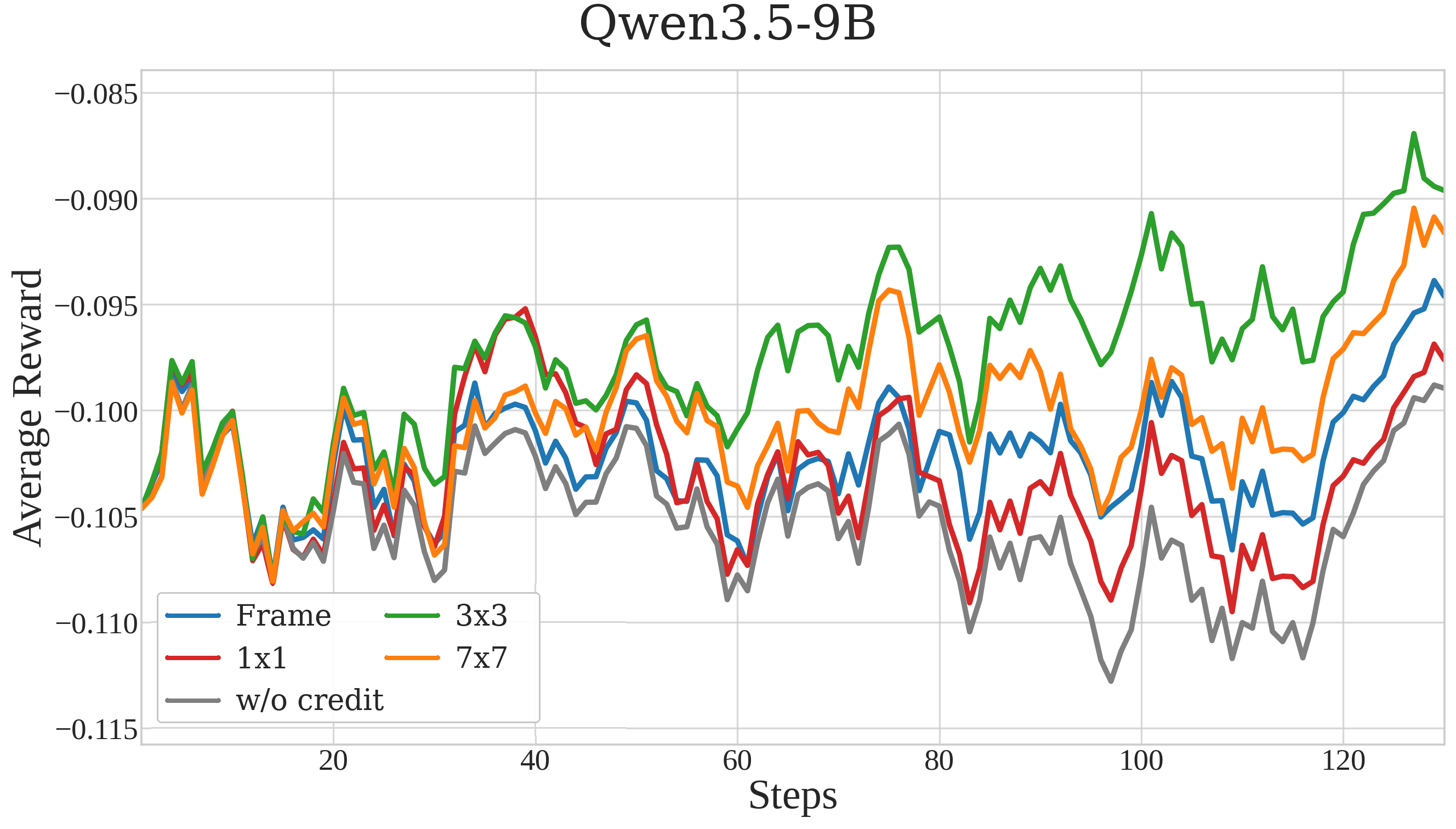}
\end{minipage}
\hfill
\begin{minipage}{0.49\textwidth}
\centering
\includegraphics[width=\linewidth]{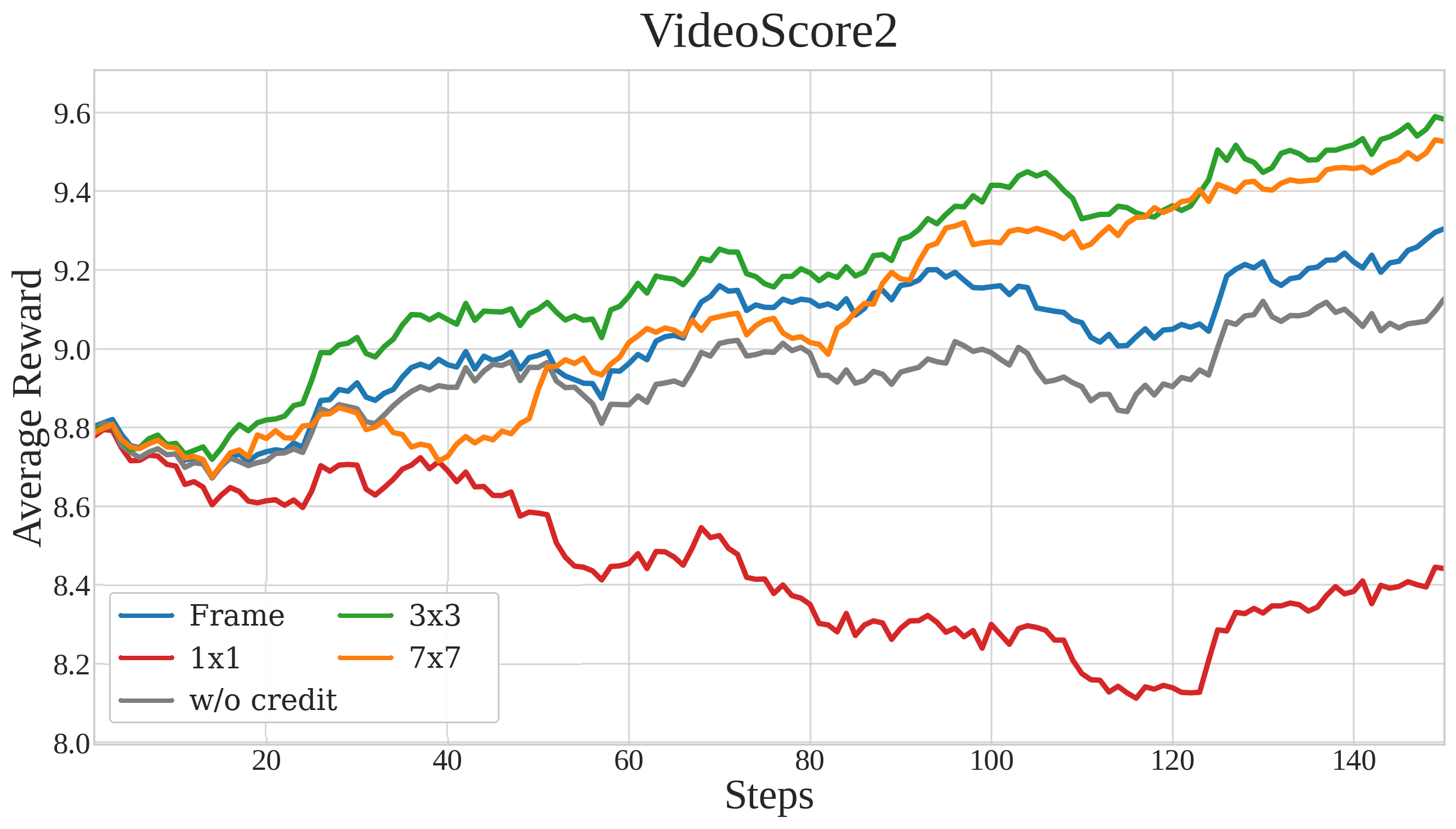}
\end{minipage}
\caption{Supplementary optimization traces including the unsmoothed $1\!\times\!1$ credit map for Qwen3.5-9B (left) and VideoScore2 (right). The cell-level variant does not continue the improvement obtained by progressively localizing spatially aggregated credit and is particularly unstable with VideoScore2.}
\label{fig:reward-curves-with-1x1}
\end{figure*}

The main-text optimization comparison focuses on the consistent w/o-credit $\rightarrow$ frame $\rightarrow$ $7\!\times\!7$ $\rightarrow$ $3\!\times\!3$ progression. \Cref{fig:reward-curves-with-1x1} additionally reports the unsmoothed $1\!\times\!1$ control. This extreme does not continue the trend: it underperforms the spatially aggregated variants with Qwen3.5-9B, and its VideoScore2 reward drops sharply, indicating evaluator-dependent optimization instability. A held-out Qwen3.5-9B evaluation is consistent with this observation: $1\!\times\!1$ obtains 55.10 Overall, below frame credit (56.47), $7\!\times\!7$ (57.08), and $3\!\times\!3$ (57.69).

The raw VLM-gradient map is a local sensitivity estimate rather than a noise-free causal attribution. Without spatial aggregation, isolated high-frequency responses can survive normalization and assign disproportionate credit to a few latent cells. This lowers the effective number of credited units and can increase update variance, as described by \cref{eq:app_credit_granularity_variance}. PPO clipping limits excessive likelihood-ratio contributions but cannot correct a persistently mislocalized routing direction. The more severe VideoScore2 failure suggests that robustness at cell-level resolution also depends on the structure of the evaluator's reward sensitivity; it implies neither that question-level advantages are required nor that token credit is ineffective.

\begin{figure}[!htb]
\centering
\includegraphics[width=1\textwidth]{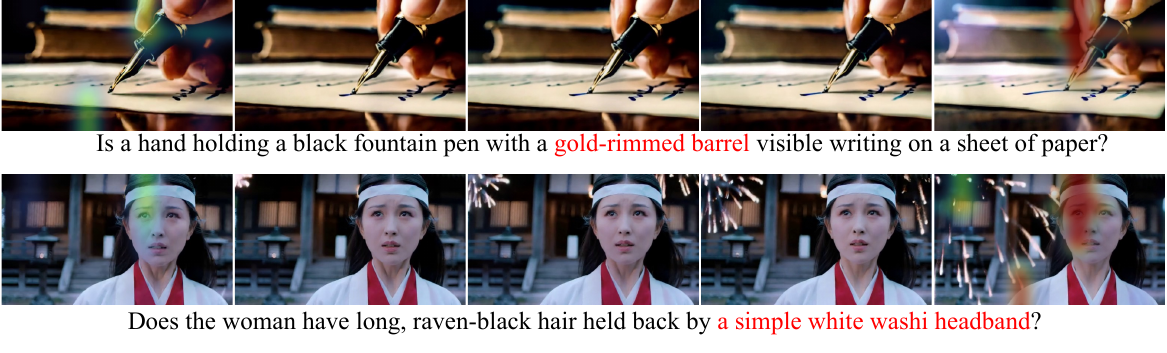}
\caption{\textbf{Additional spatial residual diagnostics.}
The positive $3\!\times\!3-7\!\times\!7$ residual for two further question-conditioned maps, shown over five sampled frames (queried evidence in red). At high-credit frames, the strongest residuals lie near the pen and gold-rimmed barrel and near the white headband. These per-example normalized sensitivity maps are not causal masks.}
\label{fig:residual-credit-appendix}
\end{figure}

Together with the held-out result above, \cref{fig:reward-curves-with-1x1} shows that removing spatial aggregation does not extend the improvement obtained by moving from frame-level to spatially aggregated credit. The complementary $3\!\times\!3$ versus $7\!\times\!7$ residual diagnostic appears in \mainref{fig:residual-credit} of the main paper, and \cref{fig:residual-credit-appendix} shows two further examples, with the strongest residuals near the pen barrel and writing region and near the white headband.

\end{document}